\documentclass[times,twocolumn,final]{elsarticle}
\usepackage{style_arxiv}
\usepackage{framed,multirow}
\usepackage{breqn}
\usepackage{siunitx}
\usepackage{amssymb}
\usepackage{latexsym}

\usepackage{url}
\usepackage{xcolor}
\usepackage{graphicx} 
\usepackage{pifont}
\usepackage{makecell}
\usepackage{enumitem}
\usepackage[titletoc]{appendix}
\usepackage{placeins}

\makeatletter
\newcommand*{\rom}[1]{\expandafter\@slowromancap\romannumeral #1@}
\makeatother

\usepackage{amsmath} 
\usepackage{xcolor}

\usepackage[colorlinks=true,urlcolor=green]{hyperref}        

\definecolor{newcolor}{rgb}{.8,.349,.1}

\begin{document}

\verso{Vinkle Kumar Srivastav \textit{et~al.}}

\begin{frontmatter}

\title{FUN-SIS: a Fully UNsupervised approach for Surgical Instrument Segmentation}%

\author[1,2]{Luca \snm{Sestini}\corref{cor1}}
\cortext[cor1]{Corresponding author: }
  \ead{sestini@unistra.fr}
\author[1]{Benoit \snm{Rosa}}
\author[2]{Elena \snm{De Momi}}
\author[2]{Giancarlo \snm{Ferrigno}}
\author[1,3]{Nicolas \snm{Padoy}}

\address[1]{ICube, University of Strasbourg, CNRS, IHU Strasbourg, France}
\address[2]{Department of Electronics, Information and Bioengineering, Politecnico di Milano, Milano, Italy}
\address[3]{IHU Strasbourg, Strasbourg, France}

\begin{abstract}
Automatic surgical instrument segmentation of endoscopic images is a crucial building block of many computer-assistance applications for minimally invasive surgery. So far, state-of-the-art approaches completely rely on the availability of a ground-truth supervision signal, obtained via manual annotation, thus expensive to collect at large scale. In this paper, we present FUN-SIS, a Fully-UNsupervised approach for binary Surgical Instrument Segmentation. FUN-SIS trains a per-frame segmentation model on completely unlabelled endoscopic videos, by solely relying on implicit motion information and instrument \textit{shape-priors}. We define \textit{shape-priors} as realistic segmentation masks of the instruments, not necessarily coming from the same dataset/domain as the videos. The \textit{shape-priors} can be collected in various and convenient ways, such as \textit{recycling} existing annotations from other datasets. We leverage them as part of a novel generative-adversarial approach, allowing to perform unsupervised instrument segmentation of optical-flow images during training. We then use the obtained instrument masks as pseudo-labels in order to train a per-frame segmentation model; to this aim, we develop a \textit{learning-from-noisy-labels} architecture, designed to extract a clean supervision signal from these pseudo-labels, leveraging their peculiar noise properties. We validate the proposed contributions on three surgical datasets, including the MICCAI 2017 EndoVis Robotic Instrument Segmentation Challenge dataset. The obtained fully-unsupervised results for surgical instrument segmentation are almost on par with the ones of fully-supervised state-of-the-art approaches. This suggests the tremendous potential of the proposed method to leverage the great amount of unlabelled data produced in the context of minimally invasive surgery.
\end{abstract}

\end{frontmatter}

\section{Introduction}
\label{sec:introduction}

\begin{figure*}[!ht]
    \centering
      \includegraphics[width=7in]{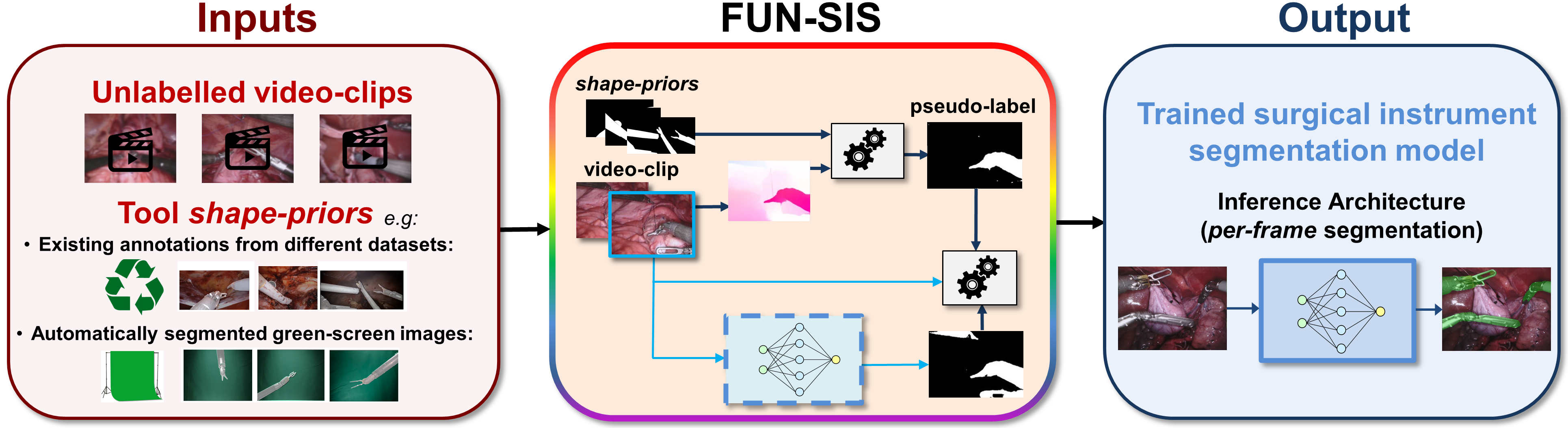}
      \caption{Paper contribution from the input-output point-of-view. The proposed FUN-SIS approach allows to train a model for surgical tool segmentation requiring as inputs only unlabelled video-clips and tool \textit{shape-priors}, obtainable in various convenient ways (e.g. by recycling existing annotations from other datasets). The method is based on a novel approach for unsupervised surgical tool segmentation of optical-flow images, generating pseudo-label masks, and a newly designed \textit{learning-from-noisy-labels} strategy, allowing to extract a clean supervision signal to train a per-frame segmentation model.}
      \label{fig:first_pic}
\end{figure*}
Minimally Invasive Surgery (MIS) has established itself as an advantageous alternative to standard open-surgery in several surgical specialties, such as pancreatic and hepatic resections (\cite{chen2018comparison}), cholecystectomy (\cite{antoniou2014meta,coccolini2015open}), appendectomy (\cite{biondi2016laparoscopic}) and inguinal hernia (\cite{takayama2020laparoscopic}). Advantages of MIS mainly derive from the small incisions through which procedures are performed, resulting in several benefits for the patients, such as reduced pain, shorter hospitalization time and less risks of infection. However, together with its benefits, MIS has also introduced new challenges for the surgeons, such as a significantly reduced field-of-view and a complex hand-eye coordination, contributing to an overall increased cognitive workload and a prolonged learning curve (\cite{harrysson2014systematic}). In order to tackle these challenges, Computer-Assistance has strongly developed in recent years, with the aim to support surgeons through a broad spectrum of applications, including automatic surgical skill analysis (\cite{zia2018automated}), surgical phases segmentation (\cite{twinanda2016endonet}), tool-tissue interaction estimation (\cite{nwoye2020recognition}), surgical scene reconstruction (\cite{long2021dssr}), field-of-view expansion (\cite{bano2020deep}), safety checkpoint evaluation~ (\cite{mascagni2021artificial}).
For most of these high-level tasks, a crucial building-block is represented by the precise localization of surgical tools in the image space, mainly by  pixel-wise classification (i.e. image segmentation).\\
State-of-the-art approaches for surgical tool segmentation use Deep Learning in order to learn a direct and general mapping between input frames and segmentation masks, robust to challenging factors such as motion blur, occlusions, cluttered background and varying lighting conditions. However, despite the unprecedented results provided by Deep Learning, the problem is still far from being solved for real-world applications: current state-of-the-art Deep Learning approaches rely heavily on manual annotations, which are expensive to obtain at a scale large-enough to allow generalization to real-world scenarios.\\ Alternatives to standard \textit{in-house annotate \& train} pipelines have been proposed, trying to address the annotation problem by cutting the cost of labels, for example by acquiring them through crowd-sourcing platforms (\cite{maier2016crowd}) or by generating semi-synthetic datasets with automatically obtained labels (\cite{garcia2021image}). General object segmentation has been tackled in an unsupervised way when video data are available, such as in Video Object Segmentation (VOS), mainly by leveraging the hypothesis of incoherent background motion, uncorrelated with the foreground (\cite{yang2019unsupervised}). However, state-of-the-art VOS approaches, as they strongly rely on such an hypothesis, tend to fail in the surgical scenario, where foreground (surgical tools) and background (tissue) strongly interact, resulting in coherent and correlated motion. 

\subsection{Contribution}
In this paper, we present FUN-SIS, a novel Fully-UNsupervised approach for binary Surgical Instrument Segmentation. The proposed approach allows to effectively train a binary surgical tool segmentation model on completely unlabelled endoscopic videos, solely relying on implicit motion information and instrument \textit{shape-priors}. We define \textit{shape-priors} as binary segmentation masks of the target object, not necessarily coming from the same dataset/domain as the videos. In the specific case of surgical tool segmentation, \textit{shape-priors} can be obtained in convenient and various ways, such as projecting 3D virtual/CAD model of  instrument on the image-space, automatically segmenting green-screen recordings, or using existing annotations from other datasets (Figure \ref{fig:first_pic}). \\
In order to achieve this, we make the following contributions:
\begin{itemize}

    \item we propose a new \textit{generative-adversarial} approach for surgical tool segmentation of optical-flow images, based on simultaneous generation and segmentation of optical-flow images from the \textit{shape-priors}. Compared to state-of-the-art Video Object Segmentation approaches, we relax the hypothesis of incoherent background motion, generally not verified in the surgical domain, letting the \textit{generative-adversarial} training process adapt to the domain characteristics. This leads to state-of-the-art results both on surgical and general Video Object Segmentation datasets;

    \item we extensively investigate the noise properties of the segmentation masks generated using the proposed optical-flow segmentation approach (\textit{pseudo-labels}), and their impact on neural-network training. We identify and thoroughly analyze two notable properties, namely \textit{unpredictability} and \textit{polarization}, and show that they can be exploited to largely improve segmentation results;
    
    \item we propose a novel \textit{learning-from-noisy-labels} strategy, based on an extended \textit{Teacher-Student} approach, allowing to train a \textit{Student} model only on \textit{probably} well-labelled regions of the noisy pseudo-labels. Differently from existing approaches, usually requiring a \textit{Teacher} model trained on clean labels, we carry out an efficient region selection in a fully-unsupervised way, exploiting the aforementioned noise properties. The proposed approach leads to high-quality segmentation results on several surgical datasets, including the popular EndoVis 2017 Instrument Segmentation dataset, while not requiring any ground-truth annotation for the training data.

\end{itemize}

\section{Related Work}

\subsection{Surgical Tool Segmentation}
Surgical tool segmentation is the task of labelling each pixel of an image as belonging to a specific class among \textit{background} and \textit{tool}, using a single \textit{instrument} class in case of binary segmentation.
First attempts to solve this problem used hand-crafted image features, machine learning models (Support Vector Machine, Random Forests) and template matching using tool \textit{shape-priors} (\cite{bouget2015detecting}, \cite{rieke2016real}). Nowadays, research works mostly address the problem of surgical tool segmentation using fully-supervised Deep Learning approaches, which have proved to outperform other existing methods (\cite{bodenstedt2018comparative}). In particular, Convolutional Neural Network (CNN) architectures have been widely adopted. \cite{garcia2017toolnet} propose a multi-scale and holistically nested CNN light-weight architecture trained with Dice loss function; \cite{shvets2018automatic} modify a VGG16 architecture by adding skip connections, winning the 2017 MICCAI EndoVis Robotic Instrument Segmentation challenge (\cite{allan20192017}). \cite{pakhomov2019deep} leverage deep residual learning and dilated convolutions for binary and part tool segmentation. \cite{hasan2019u} propose a variation of the standard U-Net architecture (\cite{ronneberger2015u}) having a modified decoder and an improved augmentation pipeline. Multi-task learning has also been explored by \cite{laina2017concurrent}, by simultaneously learning segmentation and tool landmarks localization, and by \cite{islam2021st}, by introducing the Spatio-Temporal Multi-Task Learning (ST-MTL) model, for surgical instrument segmentation and task-oriented saliency detection. \cite{jin2019incorporating} propose an attention based approach leveraging motion information, in the form of optical-flow, to improve segmentation accuracy. \cite{ni2020barnet} propose a bilinear attention network with an adaptive receptive field to tackle the challenges of scale and illumination inter-frames variability. \cite{kurmann2021mask} propose an alternative to standard semantic segmentation, first extracting instrument instances and then independently classifying them, reaching state-of-the-art results for this task. Despite the good results obtained by fully-supervised methods, their application is inherently limited by the need for manual annotations, which prevents their scalability. In order to mitigate this problem, \cite{garcia2021image} produce semi-synthetic samples, merging automatically segmented tools from green-screen recordings and real surgical background images. In the context of robotic surgery, \cite{colleoni2020synthetic} propose the combined use of recorded kinematics and green-screen, in order to cheaply obtain ground-truth segmentation masks. Several works have also tried to tackle the segmentation problem by including synthetic or unlabelled data, in combination with generative approaches. \cite{sahu2020endo} propose Endo-Sim2Real, a consistency-based framework for joint training from simulated and unlabelled real data. \cite{colleoni2021robotic} propose a cycle Generative Adversarial Network (cycle-GAN) approach to convert synthetic tools into real-looking ones, to be then blended with surgical background images, to form semi-synthetic samples.
\cite{ross2018exploiting} pre-train a CNN on unlabelled data, by means of a pretext task carried out using a cycle-GAN architecture, showing a significant boost in segmentation accuracy.
\cite{kalia2021co} incorporate unlabelled data in the training process, by mapping annotated frames to the unlabelled data domain using a cycle-GAN architecture, allowing for better generalization to the unlabelled domain. \cite{marzullo2021towards} use a \textit{pix2pix} GAN to generate synthetic surgical images from rough segmentation mask of surgical instruments and tissues. In the context of robotic surgery, \cite{pakhomov2020towards} record synchronized surgical videos and kinematic joint values and then use the letter to generate synthetic annotations, projecting the estimated tool 3D shapes, obtained via forward kinematics, onto the image space; in order to take into account the possible inaccuracy of the tool model, the segmentation problem is formulated as unpaired image-to-image translation, using a cycle-GAN architecture. An alternative proposed solution to reduce the need for manual annotations is represented by semi-supervision using label propagation. \cite{zhao2020learning} propose a flow prediction and compensation framework for semi-supervised tool segmentation, propagating low hertz annotations to unlabelled data using optical-flow. Finally, an unsupervised approach is proposed by \cite{liu2020unsupervised}, which generate tool pseudo-labels using handcrafted cues, such as color distribution, and then refine segmentation results exploiting feature correlation between adjacent video frames.\\
\indent In this work we propose a fully-unsupervised approach for surgical instrument segmentation. Differently from \cite{pakhomov2020towards}, we do not make use of synchronized kinematic information, making the approach applicable to non-robotic domains (e.g. manual laparoscopy) and to unlabelled video-only datasets (e.g. EndoVis 2017 dataset). In addition, differently from \cite{liu2020unsupervised}, we do not rely on domain-specific handcrafted cues, making the approach more robust, flexible and easy to apply to different surgical domains.

\subsection{Video Object Segmentation}
Motion is an important information which is used by the human visual system for \textit{perceptual grouping}, the process of organizing the visual information in order to efficiently perceive and interact with the world.
In the general object segmentation framework, as well as for surgical tool segmentation, motion can be a very discriminative cue, easy to obtain from unlabelled videos by means of the available powerful optical-flow estimators. Given the relevance of motion, the computer-vision community has been constantly exploring the task of Video Object Segmentation (VOS). The two standard approaches to it are semi-supervised VOS and unsupervised VOS. Semi-supervised VOS aims to track a target, specified in the first frame of the sequence in the form of a segmentation mask,  across the following frames. Unsupervised VOS, instead, aims to separate a salient foreground object from the background. It is worth noticing that, despite its name, unsupervised VOS has often been tackled in literature by means of fully-supervised training (e.g. \cite{mahadevan2020making}): the \textit{unsupervised} attribute indicates, instead, that this family of methods does not need an initial mask of the object, as in semi-supervised VOS. Among the works which have attempted to tackle the unsupervised VOS problem without a ground-truth supervision signal, \cite{wang2017saliency} propose a geodesic distance based technique, achieving good accuracy, at the cost of high per-frame computation time; more recently, Deep Learning approaches have been proposed: \cite{yang2019unsupervised} propose an adversarial framework to train a neural-network to predict a binary segmentation mask from a frame and the corresponding optical-flow image; \cite{yang2021self} propose an auto-encoder formulation using iterative binding to predict the segmentation mask from optical-flow only. In the context of surgical VOS, the semi-supervised approach is not applicable, due to the repeated changes of instruments during a procedure, and to their motion in and out of the field of view, which would require a continuous re-identification of the objects to be tracked. To our best knowledge, our work represents the first attempt to perform unsupervised VOS of surgical tools, with no annotated ground-truth for training data. The reason for such lack of approaches may lie on the additional challenges that the surgical environment brings to the VOS problem: foreground (tools) and background (tissue) strongly interact with each other, resulting in correlated motion of the two and coherent background motion, thus violating the hypothesis of several state-of-the-art approaches for unsupervised VOS; in addition, tools are not necessarily subject to continuous motion as objects in general VOS datasets, and may remain still for long periods of time: methods relying on motion segmentation alone, such as \cite{yang2021self}, would then fail to capture the object in those sequences.\\
\begin{figure*}[!ht]
    \centering
      \includegraphics[width=\textwidth]{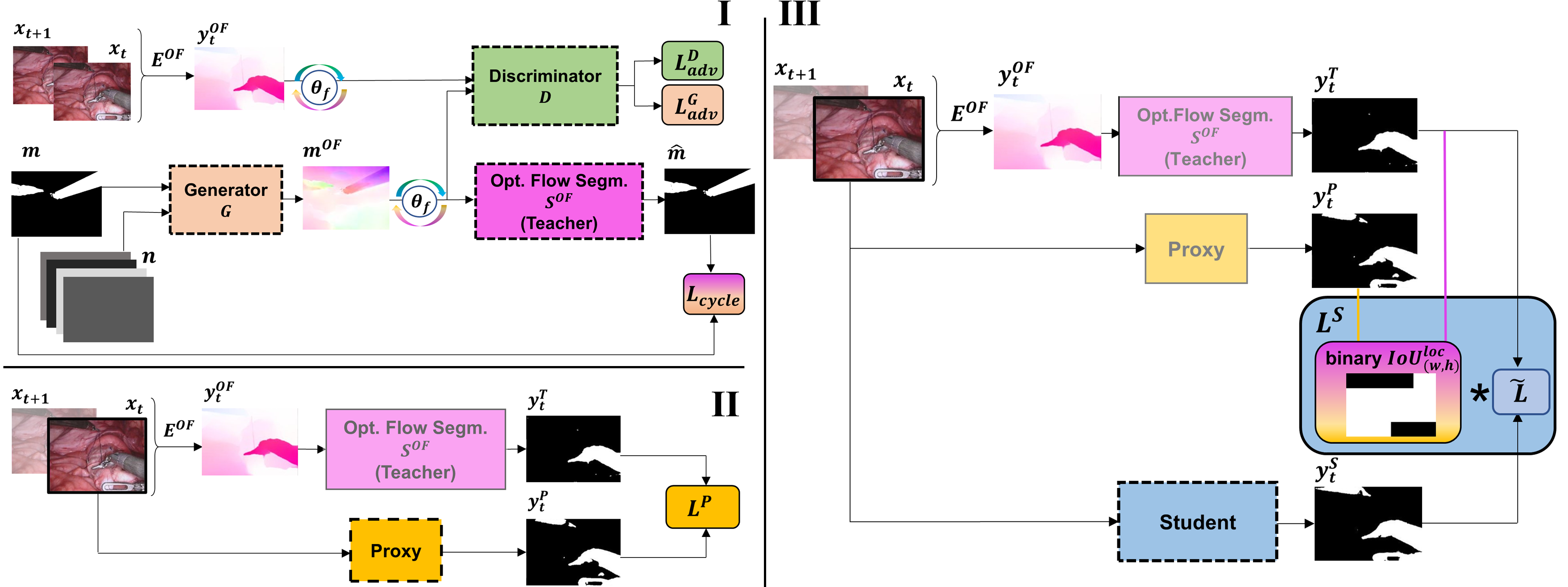}
      \caption{Overview of proposed FUN-SIS training architecture. \rom{1}: generative-adversarial training of optical-flow segmenter $S^{OF}$ (\textit{Teacher}), generator ($G$) and discriminator ($D$); generated ($m^{OF}$) and real ($E^{OF}(x_t,x_{t+1})$) optical-flow images undergo augmentation via random rotation $\theta_f$; \rom{2}: \textit{Proxy} segmentation model training, directly supervised by the pseudo-labels $y_{t}^T$, obtained from optical-flow segmentation by the \textit{Teacher} model; \rom{3}: \textit{Student} segmentation model training, leveraging \textit{local} Intersection-over-Union ($IoU_{(w,h)}^{loc}$) between \textit{Teacher} and \textit{Proxy} predictions to select well-labelled regions of $y_{t}^T$. $\tilde{L}$ is a pixel-wise loss (e.g. cross entropy), masked by the pixel-wise multiplication ($*$) with the binarized \textit{local} IoU. Loss boxes ($L$) are color coded to show which models are responsible for their minimization during training. In practice, steps \rom{1} and \rom{2} can be carried out simultaneously, as detailed in Section \ref{sec:train_strat}.}
      \label{fig:scheme}
\end{figure*}
In this work we propose a novel unsupervised approach for optical-flow tool segmentation, not requiring ground-truth annotations of the training data. In order to tackle the above mentioned challenges, we relax the hypothesis of incoherent background motion, letting a generative-adversarial training process adapt to the domain characteristics. In addition, we show that the pseudo-labels generated from optical-flow tool segmentation, even if noisy, can still provide an effective supervision signal to train a per-frame tool segmentation model, when used in synergy with an efficient \textit{learning-from-noisy-labels} strategy.

\subsection{Learning from Noisy Labels}
Effectively learning from noisy labels is becoming an essential need of Deep Learning applications.
In order to gather the massive amounts of annotations required to train Deep Learning models, researchers have recently been looking for alternatives to standard \textit{in-house} annotation, such as crowd-sourcing (\cite{yang2018leveraging}) or automatic-labelling (\cite{guo2016ms}). However, while dramatically cutting down the cost of annotations, these approaches tend to provide noisy labels. In order to tackle the \textit{learning-from-noisy-labels} problem, several approaches have been proposed in literature. Following \cite{song2020learning}, state-of-the-art approaches can be categorized in four groups. \textit{Robust Architecture} methods involve architectural modifications of standard neural networks during training, for example by adding a noise adaptation layer to model the label transition matrix of a noisy dataset (\cite{chen2015webly}). \textit{Robust Regularization} approaches involve the use of techniques such as data augmentation, weight decay, dropout, and batch normalization to prevent the overfitting of the corrupted examples. \textit{Robust Loss Design} approaches involve the modifications of standard loss functions to make them \textit{noise tolerant}. Examples include generalized cross entropy (GCE, \cite{zhang2018generalized}), symmetric cross entropy (SCE, \cite{wang2019symmetric}) and active passive loss (APL, \cite{ma2020normalized}). Finally, \textit{Sample Selection} approaches, propose strategies to select well-labelled samples. A popular approach for sample selection is multi-network training: MentorNet (\cite{jiang2018mentornet}) uses a mentor network, pre-trained on clean labels, in order to provide a curriculum for the training of a \textit{Student} network. Coteaching (\cite{han2018co}) selects \textit{probably} well-labelled samples according to a \textit{small-loss trick}, training two neural-networks in a collaborative way. While well theoretically motivated, the effectiveness of the above mentioned methods has been proven mainly in the classification task for simpler datasets than the surgical ones, such as artificially modified versions of benchmark datasets like CIFAR (\cite{lecun1998mnist}), MNIST (\cite{xiao2017fashion}) and FASHION MNIST (\cite{krizhevsky2009cifar}), and, less frequently, in real-world datasets with modest-to-medium amount of noise like ANIMAL 10-N (\cite{song2019selfie}) ($\approx 8.0\%$ noise rate), Food 101-N (\cite{lee2018cleannet}) ($\approx 18.4\%$ noise rate), WebVision (\cite{li2017webvision}) ($\approx 20.0\%$ noise rate) and Clothing 1M (\cite{xiao2015learning}) ($\approx 38.5\%$ noise rate). Segmentation differs from standard classification since pixel-labels come grouped in images. This creates the need to rethink standard methods such as \textit{Sample-Selection}, since discarding full labels may represent a waste of useful information. For this reason, local confidence map estimators have been proposed. In the context of 3D medical image segmentation, \cite{yu2019uncertainty} propose an approach for semi-supervised learning, where a segmentation model is first trained on clean labels, and then used to produce (noisy) pseudo-labels from unlabelled data, as well as confidence estimations via Monte-Carlo Dropout sampling, in order to train a \textit{Student} model only on well-labelled regions of the pseudo-labels. \cite{nie2018asdnet} also train a segmentation model on clean labels and, in parallel, a confidence model, implemented as a discriminator, in order to discriminate between predicted masks and ground-truth masks, by outputting pixel-wise scores. Unlabelled data can then be fed to segmentation and confidence models to predict pseudo-labels and local confidence maps, enriching the set of labelled training data only with high confidence regions of such predictions. While these methods have achieved good results, their effectiveness is still influenced by the amount and the quality of the available clean labels.\\
\indent In this work we tackle the problem of learning binary surgical tool segmentation from noisy pseudo-labels obtained from unsupervised segmentation of optical-flow images. Differently from the above mentioned works, our method does not require any set of clean labels in order to perform local region selection on the pseudo-labels. Instead, it leverages favorable properties of the motion-derived pseudo-labels and the finite capacity of neural-networks. These properties and the proposed method will be described in Section \ref{sec:method}.

\section{Proposed Approach}
\label{sec:method}
The FUN-SIS approach (Figure \ref{fig:scheme}) is a 3-step method which carries out unsupervised surgical tool segmentation of optical-flow images (step \rom{1}) and subsequently trains a per-frame segmentation model on the noisy pseudo-labels generated at step \rom{1} using a new \textit{learning-from-noisy-labels} strategy (steps \rom{2} and \rom{3}). The 3 steps are introduced below and detailed in the next sections:
\begin{enumerate}[label={\scshape\roman*)}]
  \item generative-adversarial training of the optical-flow tool segmentation model (called \textit{Teacher}), carried out by simultaneously learning to generate and segment synthetic optical-flow images from tool \textit{shape-priors} (Section \ref{step1});
  \item  training of a model (called \textit{Proxy}) for tool segmentation of individual frames, using, as direct supervision, the noisy pseudo-labels generated by the \textit{Teacher} model via optical-flow segmentation; the effectiveness of this step is guaranteed by a property of the noise affecting the pseudo-labels, called \textit{unpredictability} (Section \ref{step2});
  \item training of a model (called \textit{Student}) for tool segmentation of individual frames, using, as supervision, only \textit{probably} well-labelled regions of the pseudo-labels, selected according to the local agreement between the \textit{Teacher} and \textit{Proxy} models; the effectiveness of this step is guaranteed by another property of the noise affecting the pseudo-labels, called \textit{polarization} (Section \ref{step3}).
\end{enumerate}

\subsection{Step \rom{1}, Teacher: unsupervised optical-flow segmentation}
\label{step1}
The proposed approach for unsupervised optical flow-segmentation is based on a generative-adversarial approach, constrained by a cycle-consistency loss. This approach allows to learn the mapping between the domain of optical-flow images and the domain of \textit{shape-priors}, consisting of realistic binary segmentation masks of the target object (in this case surgical tools), without requiring pairwise matching between the two domains.
The method is inspired by the classic cycle-GAN architecture (\cite{zhu2017unpaired}), a popular generative architecture for image-to-image translation from unpaired domains. However, it is known that mapping between a domain of minimal complexity, as the binary \textit{shape-priors}, lacking of strong discriminative features, and a more complex one, such as the optical-flow, is an ill-posed problem, suffering from issues such as information-hiding (\textit{`steganography'} \cite{chu2017cyclegan}) and overpowering discriminator, possibly hindering the whole training process.\\
In order to deal with this \textit{complexity-imbalance}, we propose the following modifications to the standard cycle-GAN:
\begin{itemize} 
    \item we use a single cycle-consistency loss (only for \textit{shape-priors} domain), in order to avoid reconstructing a high-complexity domain sample from a \textit{synthetic} low-complexity domain sample, preventing \textit{`steganography'};
    \item we concatenate the \textit{shape-priors} domain samples with a random noise vector before feeding them to the generator. This allows the generator to produce different \textit{synthetic} optical-flow images from the same \textit{shape-priors} mask, disentangling the tool silhouette from its motion;
    \item we make intensive use of on-the-fly image augmentation.
\end{itemize}
The architecture for the proposed optical-flow segmenter is displayed in Figure \ref{fig:scheme}-\rom{1}, and discussed below.\\
Let us consider two consecutive frames belonging to a video, $x_t,x_{t+1}$ (original frames augmented by an augmentation protocol $AugmData$, consisting of random cropping and flipping), an optical-flow estimator $E^{OF}: \{x_t,x_{t+1}\} \rightarrow{y_{t}^{OF}}$, where $y_{t}^{OF}$ is the optical-flow image in the form of $[u,v]$ pixel displacement, an optical-flow generator model $G$, an optical-flow segmentation model $S^{OF}$ (also referred to as \textit{Teacher} model, due to its role in steps \rom{2} and \rom{3}), a \textit{shape-priors} binary mask $m$ and a discriminator model $D$.
The generator $G$ takes as input the \textit{shape-priors} mask $m$, augmented on-the-fly by an augmentation protocol $AugmMask$, consisting of random cropping and flipping, and concatenated with a noise vector $n$, sampled from a normal distribution of mean $\mu$ and standard-deviation $\sigma$, and resized to the input mask resolution, and outputs a synthetic optical-flow image $m^{OF}$, also in the form of $[u,v]$ pixel displacement. 
Both the real and synthetic optical-flow images, $y_{t}^{OF}$ and $m^{OF}$, undergo on-the-fly augmentation, based on augmentation protocol $AugmFlow$, and following normalization operations:

\begin{itemize}
	\item \textit{\textbf{AugmFlow}}: the optical-flow is multiplied by a random rotation matrix in the form:
    \begin{equation}
    R = 
    \begin{bmatrix}
    \cos\theta_{flow} & -\sin \theta_{flow} \\
    \sin \theta_{flow} & \cos \theta_{flow} \\
    \end{bmatrix},
    \end{equation}
    
	where $\theta_{flow}$ is randomly picked from a uniform distribution. This operation, performed on-the-fly, increases the variability of the optical-flow, and releases the generator from the burden to generate every possible flow direction;
	\item \textbf{normalization:} each optical-flow image is normalized by dividing it by the maximum pixel displacement $\sqrt{u^2+v^2}$ in it. This operation keeps the generated optical-flow image in a controlled range (where maximum displacement has norm equal to 1).
\end{itemize}

The synthetic optical-flow image $m^{OF}$ is then fed to the optical-flow segmentation model $S^{OF}$, which outputs the \textit{cycled} \textit{shape-priors} mask $\hat{m}$. The real and synthetic optical-flows $y_{t}^{OF}$ and $m^{OF}$ (both augmented and normalized) are fed to the discriminator $D$, which is trained to distinguish among the two. Cycle-consistency is ensured by requiring the cycled-mask $\hat{m}$ to match the input mask $m$ by means of a standard cross-entropy loss:
\begin{equation}
	L_{cycle} = -m\log(\hat{m}) - (1-m)\log(1-\hat{m}).
\end{equation}

Discriminator's outputs are used to enforce realistic appearance of $m^{OF}$ by training the discriminator $D$ and the optical-flow generator $G$ in an adversarial way. Specifically, the adversarial loss functions are defined as:

\begin{equation}
	L_{adv}^{G} = -\log(D(m^{OF})),
\end{equation}

\begin{equation}
	L_{adv}^{D} = -\log(1-D(m^{OF})) -\log(D(y_{t}^{OF})).
\end{equation}

The full architecture is trained end-to-end using a standard ADAM optimizer. The discriminator $D$ is trained to minimize $L_{adv}^{D}$, the optical-flow segmenter $S^{OF}$ is trained to minimize $L_{cycle}$, the optical-flow generator $G$ is trained to minimize the sum of $L_{adv}^{G}$ and $L_{cycle}$ :
\begin{equation}
	L^G = L_{adv}^{G} + L_{cycle}.
\end{equation}

\subsection{Step \rom{2}, Proxy \& the ``unpredictability'' noise property}
\label{step2}
The optical-flow segmentation by $S^{OF}$ (\textit{Teacher} model) is used to generate pseudo-labels for the unlabelled frames: each frame $x_t$ is paired with the \textit{Teacher}-generated pseudo-label mask $y_{t}^{T} = S^{OF}(y_{t}^{OF})$, which is used as direct supervision to train a neural-network (\textit{Proxy} model) to perform tool segmentation of individual frames (Figure \ref{fig:scheme}-\rom{2}).\\
The proposed approach to leverage the noisy pseudo-labels relies on findings from \cite{arpit2017closer}, which show that, while neural-networks are in principle capable of memorizing noisy samples, they tend to first take advantage of shared patterns across training examples, given their finite capacity. In a parallel study, \cite{rolnick2017deep} empirically confirmed, in the classification task, that neural-networks can generalize well even when trained on massively noisy data, rather than just memorizing noise, assuming that the noise on a pseudo-label is not conditioned by the corresponding input image itself. We define this condition as the \textit{\textbf{unpredictability}} property.\\
The noise affecting the pseudo-labels $y_{t}^{T}$ can be divided into two additive processes: the optical-flow estimation noise and the optical-flow segmentation noise. In both cases, the property of \textit{unpredictability} of noise affecting the pseudo-label $y_{t}^{T}$, from the single frame $x_t$, holds: 
\begin{itemize}
    \item the possible  absence of tool motion or presence of background coherent motion in the optical-flow image $y_{t}^{OF} = E^{OF}(x_t,x_{t+1})$, potential sources of $y_{t}^{T}$ noise, cannot be predicted from the individual frame $x_t$ only, but requires an additional frame ($x_{t+1}$) to be predicted;
    \item the optical-flow segmentation used to generate the pseudo-labels ($y_{t}^{T} = S^{OF}(y_{t}^{OF})$), second possible source of noise due to the inevitable sub-optimality of $S^{OF}$ model, does not involve the use of the frame $x_t$, contrarily to standard VOS approaches, where both frame and optical-flow are used to make a prediction (e.g. \cite{yang2019unsupervised}). 
\end{itemize} 

Given the \textit{unpredictability} property, we can train a neural-network (\textit{Proxy} model) to perform per-frame tool segmentation, using the noisy pseudo-labels $y_{t}^{T}$ directly as supervision signal. The \textit{Proxy} network takes as input the frame $x_t$ and outputs the segmentation mask $y_{t}^{P}$. The network is trained to minimize the loss $L^{P}$, which is the sum of the binary cross-entropy loss $L^{P}_{CE}$ and the log  Intersection-over-Union loss $L^{P}_{IoU}$, weighted by a factor $\alpha_P$:  
\begin{equation}
    L^{P}_{CE} = - y_{t}^{T}\log(y_{t}^{P}) - (1-y_{t}^{T})\log(1-y_{t}^{P}),
\end{equation}

\begin{equation}
    L^{P}_{IoU} = - \log \frac{\sum (y_{t}^{P}y_{t}^{T})}{\sum (y_{t}^{P} + y_{t}^{T} - y_{t}^{P}y_{t}^{T})},
\end{equation}

\begin{equation}
    L^{P} = \alpha_{P}L^{P}_{IoU} + (1-\alpha_{P})L^{P}_{CE}.
    \label{eq:loss_P}
\end{equation}

During training, the \textit{Proxy} network, unable to learn the noisy pattern from the pseudo-labels, tries to fit them with the \textit{easiest} compatible pattern, i.e. separating tools from tissue. In order to encourage this effect, we suggest the advantage of using a relatively small-capacity network compared to deeper ones. We experimentally investigate this aspect in our ablation studies, reported in Section \ref{sec:proxy_capacity}. However, as the training progresses and the pattern is learnt, the loss does not get further minimized, and gradient descent updates remain high, preventing convergence to an optimal solution. This short-coming is addressed and mitigated at step \rom{3} below.

\subsection{Step \rom{3}, Student \& the ``polarization'' noise property}
\label{step3}

\begin{figure}[tpb]
    \centering
      \includegraphics[width=3.in]{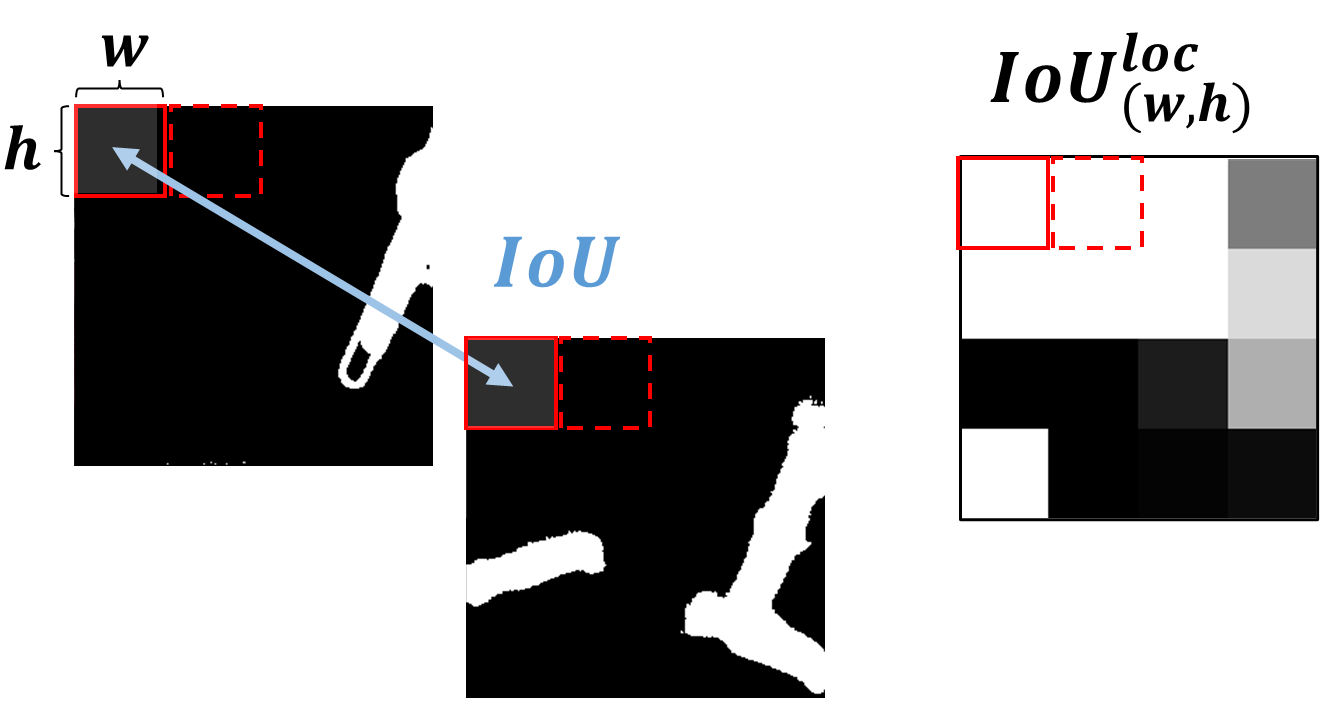}
      \caption{\textit{Local} IoU $IoU^{loc}_{(w,h)}$ is computed by sliding a window of size $w\times h$ on the two input masks, computing standard IoU at each corresponding location. The output is a single-channel image, having the same resolution as the input masks, with each pixel's value being set to the one of the IoU computed for the region it belongs to.}
      \label{fig:loc_IOU}
\end{figure}

Together with the \textit{unpredictability} property, a second peculiar property of the noise affecting the pseudo-labels $y_{t}^T$ derives from the fact that individual tools, moving coherently, tend to have a uniform appearance in the optical-flow image; this implies that, under ideal conditions (optimal optical-flow estimator $E^{OF}$, optimal optical-flow tool segmenter $S^{OF}$), each individual tool will be either perfectly segmented (if moving) or completely mislabelled (if not moving). We define the resulting noise feature as \textit{\textbf{polarization}} property, as a tool can ideally only be perfectly segmented or completely mislabelled by optical-flow segmentation. 
In the real case, this property still holds, although occlusions and sub-optimal optical-flow estimation/segmentation tend to inevitably reduce the intensity of the \textit{polarization} (i.e. there will possibly be partially segmented tools). As a practical corollary, the \textit{polarization} property suggests that inside a pseudo-label $y_{t}^T$, there will be either almost-perfectly labelled or almost-completely wrongly-labelled regions. This \textit{polarization} property will be thoroughly investigated in the experiments from Section \ref{sec:abl_noise}.\\
In order to improve training robustness and consistency, we exploit the \textit{polarization} property by designing an unsupervised method to select well-labelled regions of the pseudo-labels $y_{t}^{T}$ (Figure \ref{fig:scheme}-\rom{3}). The criterion adopted for this selection is the agreement between \textit{Proxy} network predictions $y_{t}^P$ (binarized using a threshold value $\epsilon_P$), and pseudo-labels $y_{t}^T$ (binarized using a threshold value $\epsilon_T$). The underlying idea is that the \textit{Proxy} network learns a robust general representation (the \textit{easiest} pattern). While its predictions can be incorrect at small-scale (e.g. on border pixels), they are overall reliable at greater scale (i.e. tools are not completely mislabelled as possibly happening in the pseudo-labels). In order to leverage this observation, we introduce a local version of the Intersection-over-Union (IoU) metric, called \textbf{\textit{local} IoU} ($IoU_{(w,h)}^{loc}$). In order to compute $IoU_{(w,h)}^{loc}$ between two masks, a window of size $w\times h$ is slid across the masks, using a stride equal to the window size, and IoU is computed inside each time. The output is an image with same resolution as the input masks, whose value at each pixel is the IoU computed for the region containing the pixel (Figure \ref{fig:loc_IOU}). Due to the way it is constructed, it holds that:
\begin{equation}
    \frac{1}{W\cdot H}\sum IoU^{loc}_{(W,H)} = IoU,
\end{equation}
\begin{equation}
    \frac{1}{W\cdot H}\sum IoU^{loc}_{(1,1)} = PA,
\end{equation}
where $W\times H$ is the size of the input masks, PA is the pixel accuracy metric and the summation is performed over pixels. This makes \textit{local} IoU a metric that interpolates between standard IoU and pixel accuracy, by varying the window size parameter. \textit{Local} IoU is computed between pseudo-label $y_{t}^T$ and \textit{Proxy} prediction $y_{t}^P$, and then binarized using a threshold parameter $\epsilon_{IoU}$. $\epsilon_{IoU}$ represents the minimum agreement between \textit{Proxy} and \textit{Teacher} required for a region of $y_{t}^T$ to be regarded as well-labelled. The binarized \textit{local} IoU $\overline{IoU}_{(w,h)}^{loc} = bin(IoU_{(w,h)}^{loc},\epsilon_{IoU})$ is used to prevent the loss propagation through the \textit{probably} wrongly-labelled regions of the pseudo-labels $y_{t}^{T}$, during the training of the \textit{Student} network. In particular, the \textit{Student} network takes as input the frame $x_t$ and outputs the segmentation mask $y_{t}^{S}$. The network is trained to minimize the loss $L^{S}$, which is the weighted sum of binary cross-entropy loss $L^{S}_{CE}$ and log Intersection-over-Union loss $L^{S}_{IoU}$, masked by multiplying each pixel-wise loss by $\overline{IoU}_{(w,h)}^{loc}$: 
\begin{equation}
    L^{S}_{CE} = \frac{1}{\sum \overline{IoU}_{(w,h)}^{loc}} \overline{IoU}_{(w,h)}^{loc} (-y_{t}^{T}\log(y_{t}^{S}) - (1-y_{t}^{T})\log(1-y_{t}^{S})),
\end{equation}

\begin{equation}
    L^{S}_{IoU} = -\frac{1}{\sum \overline{IoU}_{(w,h)}^{loc}} \log \frac{\sum (y_{t}^{S}y_{t}^{T}\overline{IoU}_{(w,h)}^{loc})}{\sum (y_{t}^{S} + y_{t}^{T} - y_{t}^{S}y_{t}^{T})\overline{IoU}_{(w,h)}^{loc}},
    \label{loss_iou_loc}
\end{equation}

\begin{equation}
    L^{S} = \alpha_{S}L^{S}_{IoU}  + (1-\alpha_{S})L^{S}_{CE}.
    \label{eq:loss_S}
\end{equation}


\subsection{Training Strategy}
\label{sec:train_strat}
As presented in Section \ref{sec:method} and shown in Figure \ref{fig:scheme}, the proposed approach involves a 3-step training, where the \textit{Teacher}, \textit{Proxy} and \textit{Student} models are trained successively. However, relying on the hypothesis that a neural-network will not be able to fit the noisy labels, discussed in Section \ref{step2}, we suggest that the \textit{Proxy} network can be trained on the pseudo-labels produced by \textit{Teacher} network while the \textit{Teacher} network is being trained. This allows the training to be a more compact, 2-step process, with steps \rom{1} and \rom{2} carried out simultaneously. Comparison between 3-step and 2-step training is reported in Section \ref{sec:frame_eval}.

\section{Experimental Set-Up}

\subsection{Implementation Details}
\label{sec:impl_det}
All models are implemented as neural-networks. Neural-network architectures and hyper-parameters, reported in detail in \ref{sec:app_impl_det}, were determined from preliminary experiments on external data (\textit{phantom} dataset from \cite{sestini2021kinematic}), and kept the same for all the experiments. All the segmentation models have a U-Net-like architecture. The \textit{Proxy} and \textit{Student} networks have slightly different architectures, with the \textit{Proxy} having a 11-convolutional-layer encoder (which we refer to as Unet11) and the \textit{Student} a 16-convolutional-layer (Unet16). Optical-flow estimation was carried out using RAFT (\cite{teed2020raft}), a state-of-the-art approach, trained on the publicly available non-surgical dataset FlyingThings (\cite{mayer2016large}). Training and evaluation were all carried out on $256\times256$ resized versions of the images, regardless of their original resolution/aspect ratio, due to memory constraints. The size of the noise vector $n$ was set to 32, and investigated in Section \ref{sec:abl_flowaugm}. Each value of $n$ was drawn from a normal distribution of mean $\mu$ equal to $0$ and standard-deviation $\sigma$ equal to $1$.
The $IoU_{(w,h)}^{loc}$ window size $w\times h$ was set to $64\times 64$ (1/4 of the image size); the threshold $\epsilon_{IoU}$ was set to 0.5. An in-depth study regarding $w$ and $\epsilon_{IoU}$ was carried out and reported in Section \ref{sec:abl_epsw}. The loss balancing factors $\alpha_P$, $\alpha_S$ from Equations \ref{eq:loss_P}\&\ref{eq:loss_S} were set to 0.8, and investigated in Section \ref{sec:jaccard_loss}.
Augmentations $AugmMask$ and $AugmData$ were implemented by applying random left-right, up-down flipping and random cropping, with minimal cropped region size equal to $224\times224$, then bilinearly resampled to $256\times256$. The angle $\theta_{flow}$ for the flow rotation in $AugmFlow$ was randomly picked in the range $[-\pi,\pi]$.
All augmentations were applied on-the-fly. Training was carried out using a single NVIDIA Tesla V100 GPU (32 GB). 
The code will be released upon publication.

\begin{figure*}[tpb]
    \centering
      \includegraphics[width=5.8in]{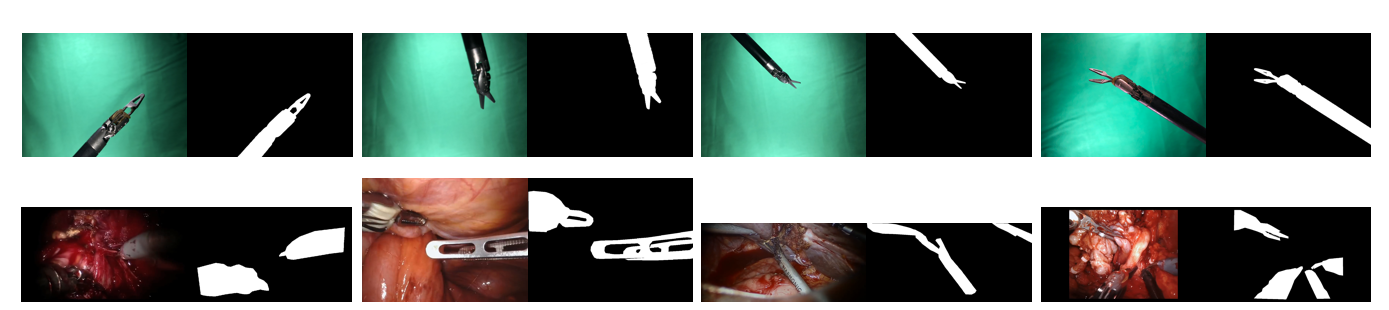}
      \caption{Examples of \textit{shape-priors} used for the EndoVis2017 experiments, and corresponding source image. Top: tools recorded in front of the green-screen and automatically segmented (\cite{garcia2021image}), called GrScreenTool; Bottom: frames from multiple robotic-assisted laparoscopic surgeries, manually segmented as part of the RoboTool dataset (\cite{garcia2021image}). Frames (and also masks) in this dataset come with various resolution/aspect ratios. Note how the appearance of the two domains is different: this is mainly due to the fact that GrScreenTool dataset, recorded using an external camera, show a different point of view on the instruments with respect to the standard surgical camera.}
      \label{fig:shape_priors_ex}
\end{figure*}

\subsection{Datasets}
\label{sec:exp_modality}
In order to validate the proposed contributions, extensive experiments were carried out, both on surgical and general object segmentation datasets.
All the data used in our experiments are now presented and categorized as \textit{Video} and \textit{Shape-priors}. Details about their use in the experiments are also reported.\\

\textbf{Video data}:
\begin{itemize}
    \item \textbf{EndoVis2017} (\cite{allan20192017}): dataset from the 2017 MICCAI EndoVis Robotic Instrument Segmentation Challenge. The dataset contains 10 video clips of abdominal porcine procedures, performed using da Vinci Xi systems. Each video contains a total of 300 high-resolution frames (1280 × 1024), recorded at 2 Hz. In the challenge 8x 225 frames were used for training, while the remaining 8x 75 frames and another 2x 300 frames were held out by the organizers for testing. According to the challenge rules, man-made devices not belonging to the da Vinci system (e.g. drop-in Ultra-Sound probe), labelled by the organizers as part of a class called \textit{Other}, are to be included in the \textit{background} class for the binary segmentation task. This introduces the need for a model to perform a semantic differentiation inside the \textit{instrument} class (da Vinci instruments and \textit{Other} instruments), which goes beyond the scope of motion-based segmenters. For this reason we refer to the dataset labelled according to the challenge rules as \textbf{EndoVis2017Challenge}, and also consider a second version of it, called \textbf{EndoVis2017VOS}, where both da Vinci and other man-made devices are labelled as \textit{instrument}. For the main experiments, we report results on both.\\
    We provide results on this dataset according to 2 modalities: 1) following the same evaluation protocol as \cite{shvets2018automatic}, by performing 4-fold cross-validation on the 8x 225 released training data (regrouped in 4 splits), and reporting the average metric on the 4 splits, for direct and fair comparison with other state-of-the-art approaches; 2) by training on RandSurg, a dataset of unlabelled data, described below, and testing on the 8x 225 EndoVis2017VOS frames.
    \item \textbf{RandSurg}: this dataset consists of 4 full unlabelled laparoscopic robotic-assisted procedures downloaded from a public repository (\cite{videos}): adhesiolysis (1036 frames), inguinal hernia repair (1075 frames), appendectomy (500 frames) and ex-vivo suturing demo (525 frames).  A set of experiments was carried out by training our model on this dataset and evaluating the performance on EndoVis2017VOS; in order to simulate a realistic application of the FUN-SIS method, and show its ease-of-use, the videos underwent minimal pre-processing (cropping, no trimming, so possibly including  out-of-body scenes).
    \item \textbf{STRAS}: this dataset is obtained from endoscopic submucosal dissection procedures performed through the STRAS robotic system (\cite{de2013introducing}), a robotic system consisting of a robotized endoscope, having two lateral channels for flexible robotic tools. The dataset was built from a 5 days-experiment on porcine models\footnote{The study protocol for this experiment was approved by the Institutional Ethical Committee on Animal Experimentation (ICOMETH No.38.2011.01.018). Animals were managed in accordance with French laws for animal use and care as well as with the European Community Council directive no. 2010/63/EU} (\cite{zorn2017novel}), recorded at 30 fps. Each frame was paired with another 1 second apart in the future, for optical-flow computation. The whole dataset was resampled regularly, yielding a total of 5644 frames ($\sim$1100 per experiment day). For each day, 200 frames, regularly spaced, were manually annotated for evaluation (1000 annotated samples in total). The dataset contains challenging sequences, involving bleeding, smoke, strong tool-tissue interaction and image blurring. We provide results on this dataset by performing 5-fold cross-validation (each fold corresponding to an experiment day), and reporting the average metric on the 5 splits.
    \item \textbf{Cholec80} (\cite{twinanda2016endonet}): dataset containing 80 unlabelled videos of manual laparoscopic cholecystectomy  procedures captured at 25 Hz and resampled at 1 Hz. We provide qualitative results on this dataset by using the standard split (40 videos for training, 40 videos for testing) to show cross-surgery applicability of the proposed FUN-SIS method.
    \item \textbf{DAVIS2016} (\cite{perazzi2016benchmark}): a popular VOS dataset, containing different moving objects (e.g. animals, people, cars). The dataset consists of 50 clips for a total of 3455 1080p frames with pixel-wise annotations. We provide results on this dataset in order to evaluate the proposed optical-flow segmentation approach on non-surgical videos. To this aim, the standard training-test split was used (30 videos for training and 20 for testing), for fair comparison with  state-of-the-art VOS approaches.
\end{itemize}

\textbf{Shape-priors}:
\begin{itemize}
    \item \textbf{RoboTool}: 514 manually segmented tool masks, from the RoboTool dataset, released by \cite{garcia2021image}. Examples of the original frames and manually segmented tools can be see in Figure \ref{fig:shape_priors_ex}, bottom. Original masks were cropped to remove the lateral black bands, and resized to $256\times256$ regardless of their original aspect ratio.
    \item \textbf{GrScreenTool}: automatically segmented tools from recordings in front of a green-screen. A total number of 1100 masks were downloaded from the publicly released dataset by \cite{garcia2021image}, mostly having a single tool. Random couples of masks were then selected and merged together, in order to avoid having single-tool masks. Following this strategy, a total number 2200 masks were obtained. Examples of the original green-screen images and extracted tools can be seen in Figure \ref{fig:shape_priors_ex}, top.
    \item \textbf{STRASMasks}: 2000 projections of approximate 3D virtual/CAD model of the two STRAS tools, used as \textit{shape-priors} in the STRAS experiments; Details regarding the projection operation can be found in \cite{sestini2021kinematic}.
    \item \textbf{SegTrackV2} (\cite{li2013video}): 976 manual annotations from the generic VOS dataset SegTrackV2. The dataset includes different segmented objects (e.g. animals, cars, people), used as \textit{shape-priors} in the DAVIS2016 experiments.
    \item \textbf{FBMS59} (\cite{ochs2013segmentation}):
    720 manual annotations from the generic VOS dataset FBMS59. The dataset includes different segmented objects (e.g. animals, cars, people), used as \textit{shape-priors} in the DAVIS2016 experiments.
    
\end{itemize}

\begin{figure}[tpb]
    \centering
      \includegraphics[width=3.5in]{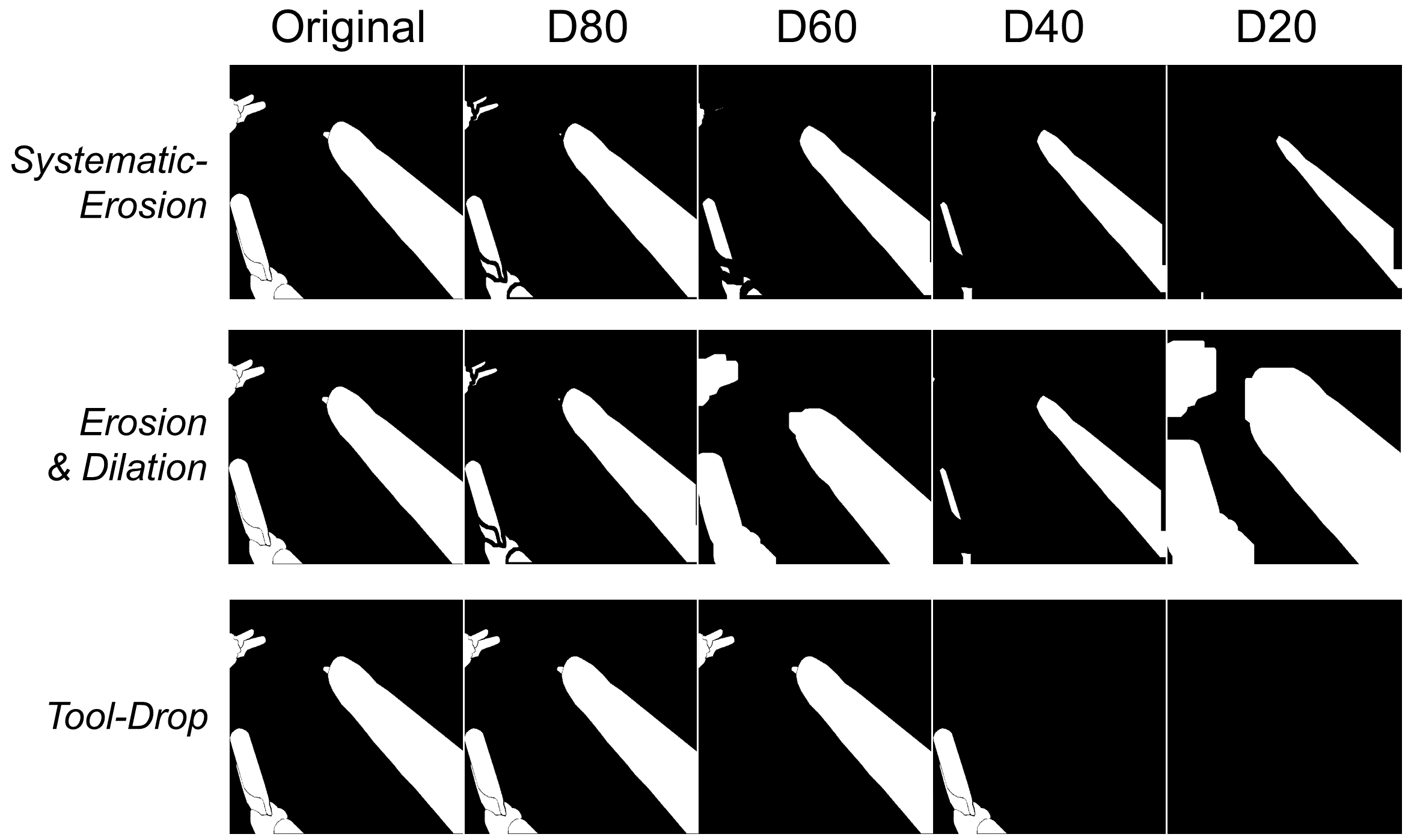}
      \caption{Samples from the artificially-corrupted versions of EndoVis2017 dataset.
        From top to bottom: \textit{Systematic Erosion}, \textit{Erosion \& Dilation}, \textit{Tool-Drop}. For each noise source, a sample from D80 ($\sim$80\% mean IoU between training sample labels and original ones), D60 ($\sim$60\% mean IoU), D40 ($\sim$40\% mean IoU), D20 ($\sim$20\% mean IoU) is shown.}
      \label{fig:masks}
\end{figure}

\begin{table*}[t]
\centering
\begin{tabular}{|l|c|c|c|}
\hline
\phantom{6} & Annot. [\%] & EndoVis2017 & DAVIS2016\\
\hline
Baseline\textsubscript{FS} & 100 & 60.47 & 73.58\\
\hline
\hline
CIS (\cite{yang2019unsupervised}) & 0* & 24.15 & 60.89 (71.5)\\
Teacher\textsubscript{RoboTool}(ours) & 0 & 40.08 & /\\
Teacher\textsubscript{GrScreenTool}(ours) & 0 & \textbf{40.47} & /\\
Teacher\textsubscript{FBMS}(ours) & 0 & / & 62.72\\
Teacher\textsubscript{SegTrackV2}(ours) & 0 & / & \textbf{63.40}\\
\hline
\end{tabular}
\caption{Optical-flow segmentation. Comparison of the proposed method (\textit{Teacher}), using different \textit{shape-priors} for training (RoboTool, GrScreenTool for EndoVis2017VOS experiments; FBMS, SegTrackV2 for DAVIS2016 experiments), with the state-of-the-art CIS approach (without and with post-processing, in parenthesis) and a fully-supervised baseline (Baseline\textsubscript{FS}). Mean IoU [\%] is reported. Percentage of annotated training samples required by each method is also reported (Annot. [\%]). Note that CIS uses frames and optical-flow to make predictions, while our approach only uses optical-flow.}
\label{tab:flow}
\end{table*}

\subsection{Artificially Corrupted dataset}
\label{sec:data_noisy}
In order to gain a full understanding of the impact of the noise properties presented in Section \ref{step2} and \ref{step3} on the proposed \textit{learning-from-noisy-labels} approach, we also perform experiments on the EndoVis2017VOS dataset under controlled noise conditions. For these experiments we substitute, in our training pipeline, the pseudo-labels $y_{t}^{T}$ generated by the \textit{Teacher} network, with artificially corrupted versions of the clean labels.
To this aim, we consider three types of label corruption, described below: 
\begin{itemize}
    \item \textit{Systematic-Erosion}: each ground-truth mask is eroded;
    \item \textit{Erosion \& Dilation}: each ground-truth mask is randomly eroded or dilated;
    \item \textit{Tool-Drop}: \textit{full} tool annotations are randomly dropped (i.e. each tool is either \textit{perfectly}-annotated or not-annotated at all).
\end{itemize}

For each noise type we apply the corresponding transformation, modulating its intensity in order to obtain 4 datasets, \{D80, D60, D40, D20\}, each one having a mean IoU between the corrupted labels and the ground-truth of $\sim$80\%, $\sim$60\%, $\sim$40\%, $\sim$20\%, respectively (e.g. greater erosion is applied to generate D20 compared to D40, in the \textit{Systematic-Erosion} experiment). Examples of the datasets are shown in Figure \ref{fig:masks}. We use this dataset as part of the ablation study detailed in Section \ref{sec:abl_noise}.

\section{Experiments and Results Analysis}  
In this section we present experimental results and comparisons with state-of-the-art methods. First, we analyze the effectiveness of the proposed optical-flow segmentation approach, both on surgical and general object-segmentation datasets. We then analyze results of surgical tool segmentation of individual frames. In order to evaluate model performance, mean Intersection-over-Union (IoU) between predictions and manually annotated ground-truth (GT) is used.

\begin{figure}[t]
    \centering
      \includegraphics[width=3.2in]{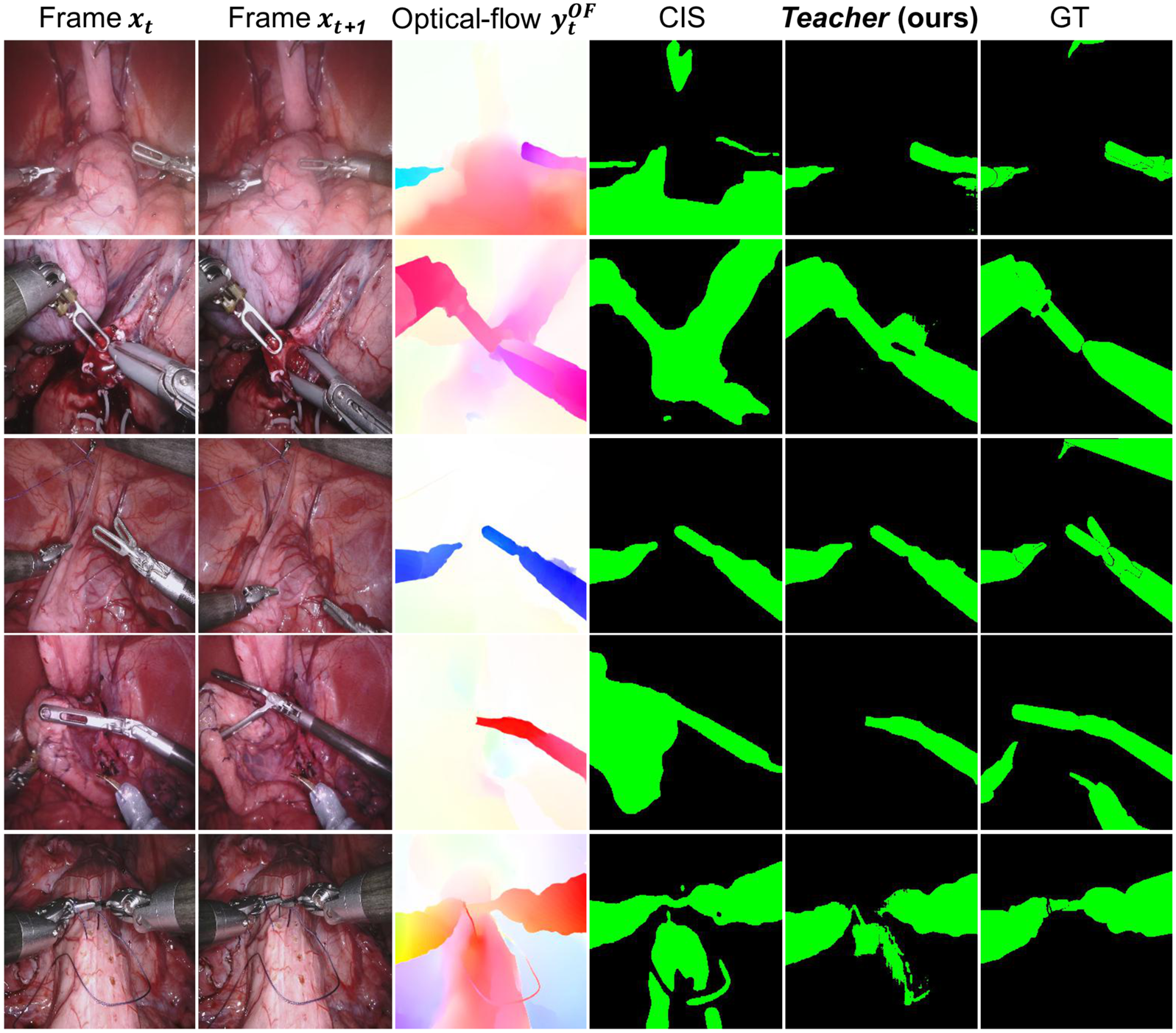}
      \caption{Optical-flow segmentation on EndoVis2017VOS. Qualitative results showing frame couples used for optical-flow computation, optical-flow images after HSV standard conversion, predictions from CIS (\cite{yang2019unsupervised}) and \textit{Teacher} (trained using RoboTool \textit{shape-priors}), and ground-truth (GT).}   
      \label{fig:qual_res_endo_flow}
\end{figure}

\subsection{Optical-Flow Segmentation}
\label{sec:exp_of}
Optical-flow segmentation by the \textit{Teacher} network was evaluated on EndoVis2017VOS and DAVIS2016, and compared with a state-of-the-art Deep Learning approach for unsupervised Video Object Segmentation, called Contextual Information Separation (CIS, \cite{yang2019unsupervised}), adopting the same evaluation protocol on DAVIS2016. We report the CIS results both with and without post-processing, for fair comparison with our approach which does not make use of it, using the trained network parameters provided by the authors for DAVIS2016 experiments. Despite being trained using the PWC-net optical-flow estimator (\cite{sun2018pwc}), we observed that the CIS model provided more accurate results using RAFT-generated optical-flow images: we thus reported results using the latter. On EndoVis2017VOS, the CIS model was trained from scratch, using the RAFT optical-flow estimator: training was carried out using the code publicly released by the authors (\cite{github_CIS}). We trained our \textit{Teacher} model using RoboTool and GrScreenTool \textit{shape-priors} for EndoVis2017VOS experiment, and SegTrackV2 and FBMS for DAVIS2016 experiment. We also report results of a fully-supervised baseline (Baseline\textsubscript{FS}) model, having the same architecture as the \textit{Teacher} network, trained on GT labels.

Experimental results, presented in Table \ref{tab:flow}, show that the proposed approach outperforms the state-of-the-art CIS approach (without post-processing) both in the surgical scenario (EndoVis2017VOS dataset) and in general object segmentation (DAVIS2016 dataset). The reason behind the significant improvement on EndoVis2017VOS (+16.32\% $\Delta$IoU) may reside in the independence of the proposed approach from the hypothesis of incoherent background motion. In fact, our method lets the generator and discriminator adapt to the complexity of the optical-flow domain, generating samples with possible cluttered background and tool occlusion, while still enforcing correct tool segmentation through the cycle-consistency loss. Examples of challenging generated optical-flow images can be seen in Figure \ref{fig:qual_gen_samples}. Deeper insights on optical-flow generation will be provided the ablation study in Section \ref{sec:abl_flowaugm}. As a result, the optical-flow segmenter becomes more robust to cluttered scenes, where tissue, as well as tools, moves coherently. As shown in the qualitative results shown in Figure \ref{fig:qual_res_endo_flow}, the proposed \textit{Teacher} model outperforms the CIS approach especially when tools interact with the anatomy (e.g. pulling tissue, second row from bottom).   

\begin{table*}[ht]
\centering
\begin{tabular}{|l|c|c|c|}
\hline
\phantom{6} & Annot. [\%] & \hphantom{Cha}EndoVis2017VOS\hphantom{nge} & EndoVis2017Challenge \\
\hline
TernausNet-16 (\cite{shvets2018automatic}) & 100  & (89.06) & 83.60 (82.95) \\
MF-TAPNet (\cite{jin2019incorporating}) & 100* & (\textbf{89.61}) & \textbf{87.56} (85.81) \\
Baseline\textsubscript{FS} & 100 & 88.99 & 82.55 
\\
\hline
\hline
AGSD (\cite{liu2020unsupervised}) & 0  & (71.47) & 67.85 (65.30) \\
Teacher (ours)& 0  & 40.08 & 37.03 \\
Proxy (ours)& 0   & 74.78 & 68.31\\
\textbf{Student (ours)}& 0  & \textbf{83.77} & \textbf{76.25} \\
\hline
\end{tabular}
\caption{Surgical tool segmentation of individual frames. Comparison of the proposed unsupervised method (trained using RoboTool \textit{shape-priors}), with state-of-the-art unsupervided AGSD approach, fully-supervised approaches TernausNet-16 and MF-TAPNet, and fully-supervised baseline (Baseline\textsubscript{FS}) on the EndoVis2017VOS and EndoVis2017Challenge datasets. Results in parenthesis for state-of-the-art approaches were obtained by training the models using the code released by the authors. Mean IoU [\%] is reported. Percentage of annotated training samples required by each method is also reported (Annot. [\%]). Note that MF-TAPNet uses 2 consecutive frames at inference time to make a prediction, while the other approaches use individual frames.}
\label{tab:all_end}
\end{table*}

\begin{table}[ht]
\centering
\begin{tabular}{|l|c|c|c|}
\hline
\phantom{6} & \textit{p-value} (t-test) & \textit{Cohen's d}\\
\hline
Proxy-Teacher & $p << 0.001$ & 1.566\\
Student-Proxy & $p << 0.001$ & 0.612\\
Baseline\textsubscript{FS}-Student & $p << 0.001$ & 0.448\\
\hline
\end{tabular}
\caption{Statistical analysis of tool segmentation results obtained in EndoVis2017VOS (Table \ref{tab:all_end}). For each pair, t-test was run (\textit{p-values} reported in first column) and \textit{Cohen's d} number was computed.}
\label{tab:stat_an}
\end{table}

\subsection{Per-frame Surgical Tool segmentation}
\label{sec:frame_eval}
Per-frame surgical tool segmentation was evaluated on the EndoVis2017Challenge, EndoVis2017VOS and STRAS datasets, according to the modalities reported in Section \ref{sec:exp_modality}, using RoboTool and STRASMasks as \textit{shape-priors}, respectively.
For each experiment, we report results for the following networks:
\begin{itemize}
    \item \textit{Teacher} network, producing the pseudo-labels $y_{t}^{T}$ from optical-flow segmentation, evaluated against GT masks;
    \item \textit{Proxy} network, directly trained on the noisy pseudo-labels, producing segmentation masks $y_{t}^{P}$ from individual frames, evaluated against GT masks;
    \item \textit{Student} network trained using \textit{local} IoU masking, producing segmentation masks $y_{t}^{S}$ from individual frames, evaluated against GT masks. The \textit{Student} network is the output model of the proposed FUN-SIS approach.
\end{itemize}

\begin{figure*}[htpb]
    \centering
      \includegraphics[width=\textwidth]{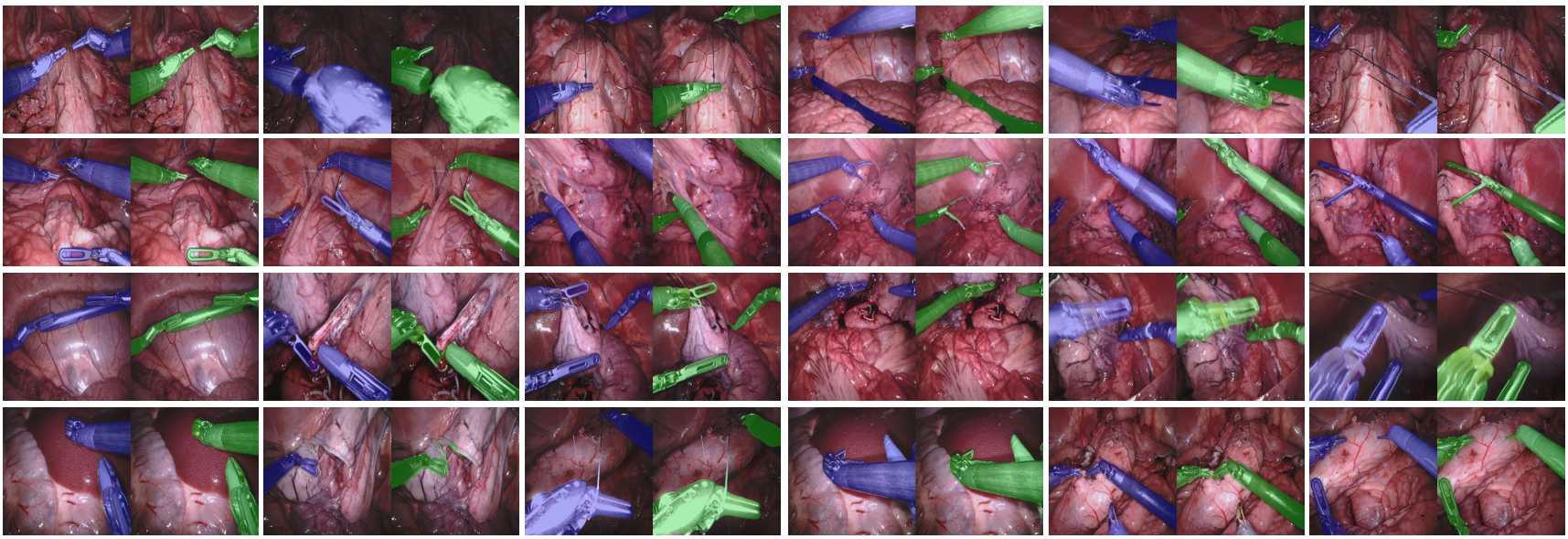}
        \caption{Qualitative results on the EndoVis2017VOS dataset, from the experiment reported in Table \ref{tab:all_end}. Original frame overlapped with ground-truth (blue) and \textit{Student} network's prediction (green).}           
        \label{fig:qual_end_add}
\end{figure*}

\begin{figure*}[htpb]
    \centering
      \includegraphics[width=\textwidth]{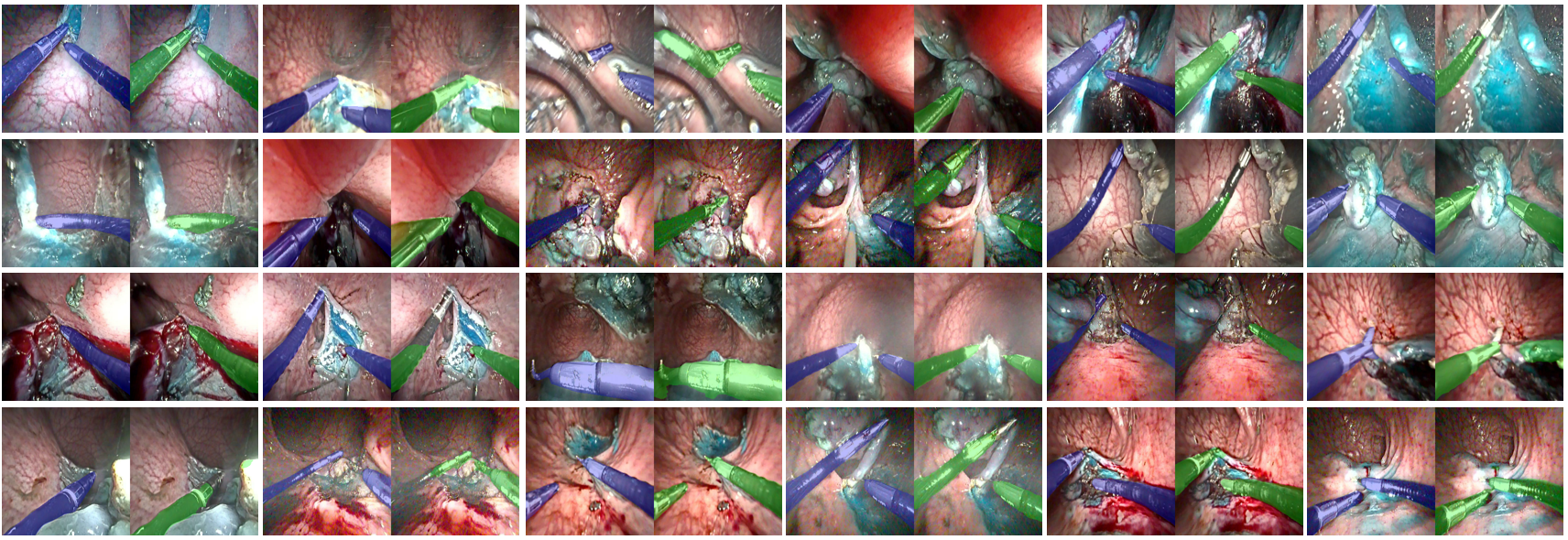}
        \caption{Qualitative results on the STRAS dataset, from the experiment reported in Table \ref{tab:all_stras}. Original frame overlapped with ground-truth (blue) and \textit{Student} network's prediction (green).}           
        \label{fig:qual_stras_add}
\end{figure*}

For the EndoVis2017Challenge and EndoVis2017VOS experiments we compare the proposed approach with the unsupervised Anchor Generation and Semantic Diffusion (AGSD) approach (\cite{liu2020unsupervised}), based on handcrafted features, and with the fully-supervised state-of-the-art approaches TernausNet-16 (\cite{shvets2018automatic}) and MF-TapNet (\cite{jin2019incorporating}). Results on EndoVis2017VOS for these approaches were obtained by training the models using the code publicly released by the authors (\cite{github_AGSD,github_ternaus,github_tapnet}). Additionally, we compare our results with Baseline\textsubscript{FS}, a model sharing the same architecture as the \textit{Student} network (Unet16), but trained in a fully-supervised way on the GT labels. We do not provide fully-supervised results on the STRAS dataset, due to the lack of GT training labels. We also do not provide results for the unsupervised AGSD approach, due to the fact that the handcrafted cues selected by the authors are specifically tailored for the EndoVis dataset, yielding poor results on the significantly different STRAS dataset.\\
\begin{figure*}[!ht]
    \centering
      \includegraphics[width=5.4in]{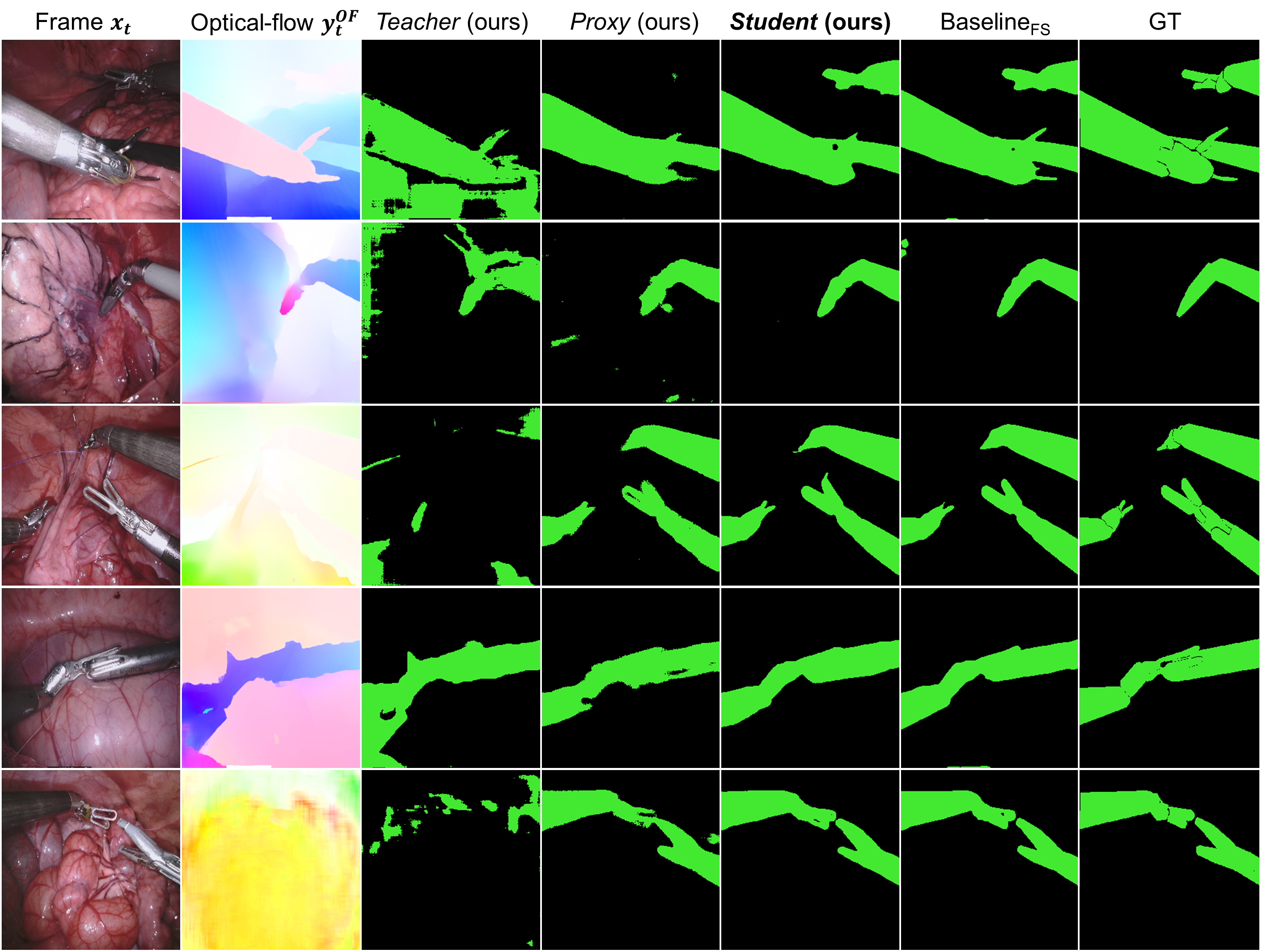}
        \caption{Surgical tool segmentation on the EndoVis2017VOS dataset. Qualitative results showing, from left to right, input frame $x_t$, optical-flow image $y_t^{OF}$ using HSV standard conversion, predictions from \textit{Teacher} (using RoboTool \textit{shape-priors}), \textit{Proxy}, \textit{Student} and fully-supervised baseline (Baseline\textsubscript{FS}), and ground-truth (GT).}  
        \label{fig:qual_res_end}
\end{figure*}
\begin{figure*}[!ht]
    \centering
      \includegraphics[width=4.7in]{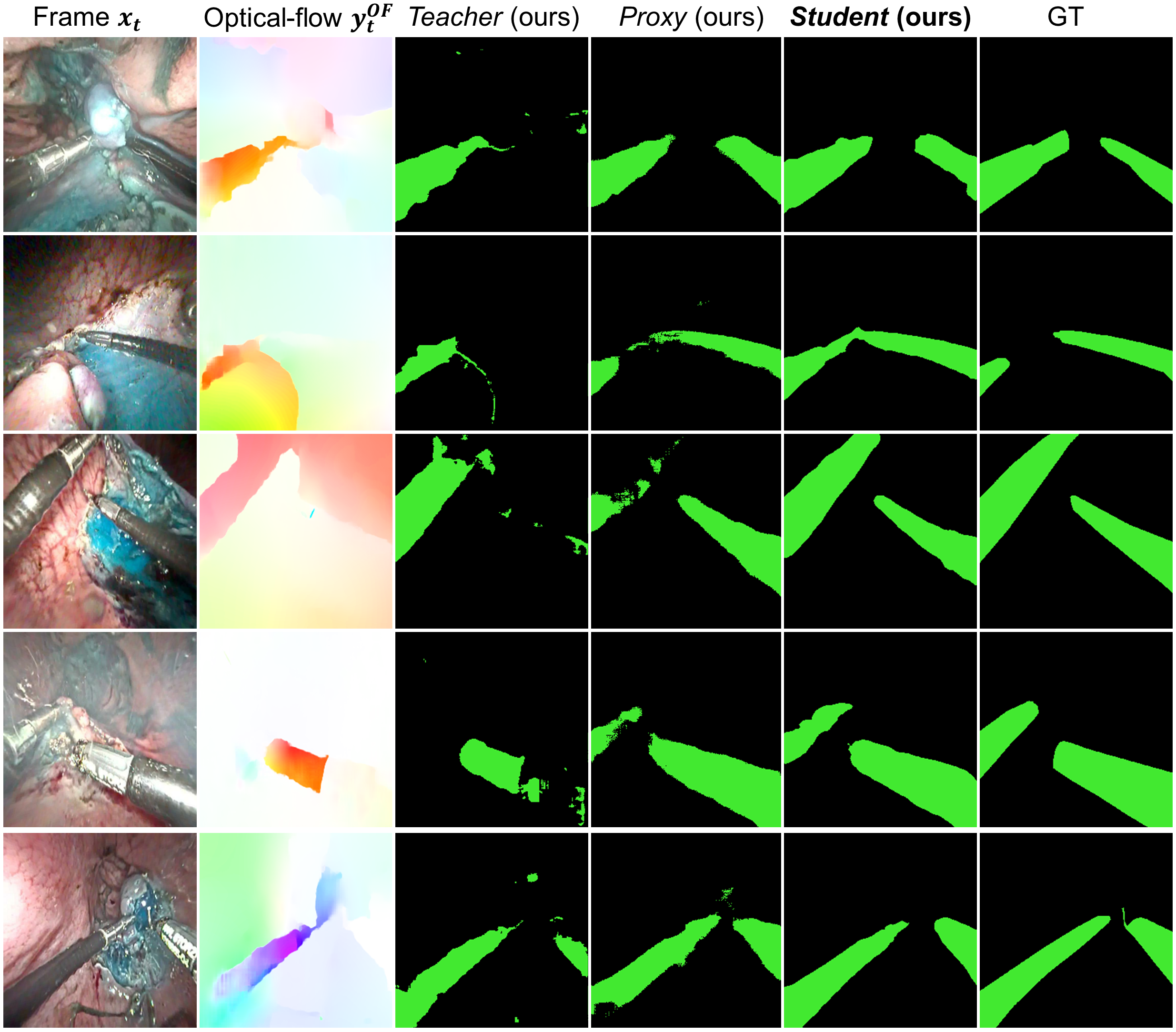}
        \caption{Surgical tool segmentation on the STRAS dataset. Qualitative results showing, from left to right, input frame $x_t$, optical-flow image $y_t^{OF}$ using HSV standard conversion, predictions from \textit{Teacher} (using STRASMasks \textit{shape-priors}), \textit{Proxy} and \textit{Student}, and ground-truth (GT).}           
        \label{fig:qual_res_stras}
\end{figure*}
\begin{figure}[t]
    \centering
      \includegraphics[width=3.3in]{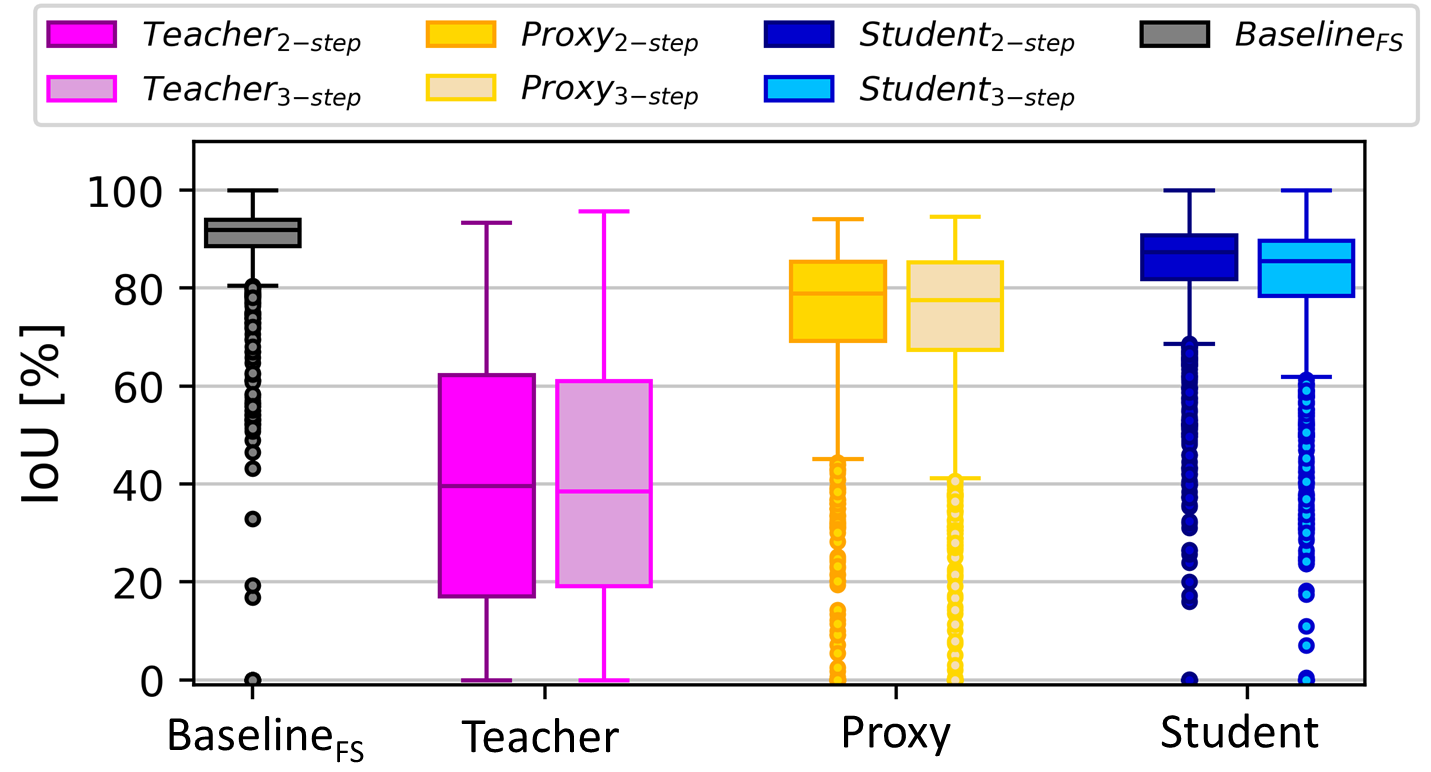}
      \caption{Box-plots showing IoU distributions from EndoVis2017VOS segmentation experiment (Table \ref{tab:all_end}). Fully-supervised baseline Baseline\textsubscript{FS} (grey), \textit{Teacher} (purple 2-step, light purple 3-step), \textit{Proxy} (yellow 2-step, light yellow 3-step), \textit{Student} (blue 2-step, light blue 3-step).}
      \label{fig:bp_end}
\end{figure}
\noindent
Experimental results, reported in Table \ref{tab:all_end}, show that the proposed approach enables to effectively train the \textit{Student} network in a fully-unsupervised way, reaching 83.77\% IoU on the EndoVis2017VOS dataset, 12.30\% above the unsupervised AGSD approach and only 5.22\% below the fully-supervised baseline. As hypothesized, the noise affecting the pseudo-labels generated by optical-flow segmentation cannot be predicted from the individual frames, thus cannot be learnt by the \textit{Proxy} network. This results in a significant improvement of the \textit{Proxy} network's predictions compared to pseudo-labels used for its training  (+34.70\% $\Delta$IoU on EndoVis2017VOS). On top of this, the \textit{Student} network significantly improves the segmentation quality, by training only on the \textit{probably} well-labelled regions of the pseudo-labels, selected by means of the \textit{local} IoU between pseudo-labels and \textit{Proxy} predictions: the improvement of the \textit{Student} network, with respect to the \textit{Proxy} network, amounts to +8.99\% $\Delta$IoU on EndoVis2017VOS. Qualitative results presented in Figure \ref{fig:qual_res_end} clearly show the dramatic improvement of the \textit{Proxy} network compared to the \textit{Teacher} network, and the refining effect of the \textit{Student} network, producing accurate and sharp segmentation masks. In order to assess statistical significance of the results on the EndoVis2017VOS dataset, pairwise t-tests were run (sample size N=1800) between \textit{Proxy} \& \textit{Teacher}, \textit{Student} \& \textit{Proxy} and \textit{baseline\textsubscript{FS}} \& \textit{Student}, all showing statistically significant differences ($p << 0.001$ for all the three pairs). In addition, \textit{Cohen's d} number was computed for such pairs, in order to quantify the strength of such statistically significant difference. \textit{Cohen's d} numbers analysis, reported in Table \ref{tab:stat_an}, shows that the effect-size of such differences is \textit{very large} between \textit{Proxy} \& \textit{Teacher} ($d > 1.2$, \textit{d} = 1.566), \textit{medium/high} between Student \& Proxy ($0.5 < d < 0.8$, \textit{d} = 0.612) and \textit{medium/small} between \textit{fully-supervised baseline} \& \textit{Student} ($0.2 < d < 0.5$, \textit{d} = 0.448) (according to \cite{cohen2013statistical,sawilowsky2009new}). As expected, the performance on the EndoVis2017Challenge dataset, where devices such as the Ultra-Sound probe are considered as part of the \textit{background} class, is lower than the one on EndoVis2017VOS, while still outperforming the unsupervised AGSD approach (+8.40\% $\Delta$IoU). This is due to the fact that our approach, despite not being trained using specific \textit{shape-priors} of these tools, is still able to generalize and segment them together with the da Vinci ones. Examples of frames containing the drop-in Ultra-Sound probe are shown in Figure \ref{fig:qual_res_end}, first and fourth row from the top. In order for our approach to learn such semantic discrimination  between the two \textit{instrument} classes, pure motion information may not be sufficient. The possible extension of FUN-SIS to multi-class segmentation will be discussed in Section \ref{sec:future}. We also analyze the difference between the 2-step and 3-step training strategies described in Section \ref{sec:train_strat}. Results, shown in Figure \ref{fig:bp_end}, confirm that the two modalities provide comparable results, as suggested in Section \ref{sec:train_strat}. We thus consider the 2-step approach superior, due to the shorter training time required. Results obtained on the challenging STRAS dataset, reported in Table \ref{tab:all_stras}, confirm the ability of the method to effectively learn surgical tool segmentation in a fully-unsupervised way. The \textit{Student} network, trained without any domain-specific hyper-parameter tuning, reaches an IoU equal to 66.37\%, despite being trained on very low-quality pseudo-labels (29.93\% IoU). As observable from Figure \ref{fig:qual_res_stras}, in fact, optical-flow images appear less sharp compared to the EndoVis2017 ones, mainly due to image blurring and lower image resolution, influencing the overall performance. The implications of the method's dependency on optical-flow quality will be discussed in Section \ref{sec:future}. Additional qualitative results for the \textit{Student} network on the EndoVis2017VOS and STRAS datasets are displayed in Figures \ref{fig:qual_end_add}\&\ref{fig:qual_stras_add}, at the end of the manuscript.

\begin{table}[t]
\centering
\begin{tabular}{|l|c|c|}
\hline
\phantom{6} & Annot. [\%] & STRAS\\
\hline
Teacher (ours)& 0 & 29.93\\
Proxy (ours)& 0 & 55.07\\
\textbf{Student (ours)}& 0 & \textbf{66.37}\\
\hline
\end{tabular}
\caption{Surgical tool segmentation of individual frames. Results of the proposed method on the STRAS dataset using STRASMasks \textit{shape-priors}. Mean IoU [\%] is reported. Percentage of annotated training samples required by each method is also reported (Annot. [\%]).}
\label{tab:all_stras}
\end{table}
\section{Ablation Studies and Additional Experiments}
\label{sec:exp_abl}
In order to provide a more in-depth understanding of the proposed FUN-SIS approach, we performed several ablation studies on crucial aspects of the method.

\begin{figure}[tpb]
    \centering
      \includegraphics[width=2.7in]{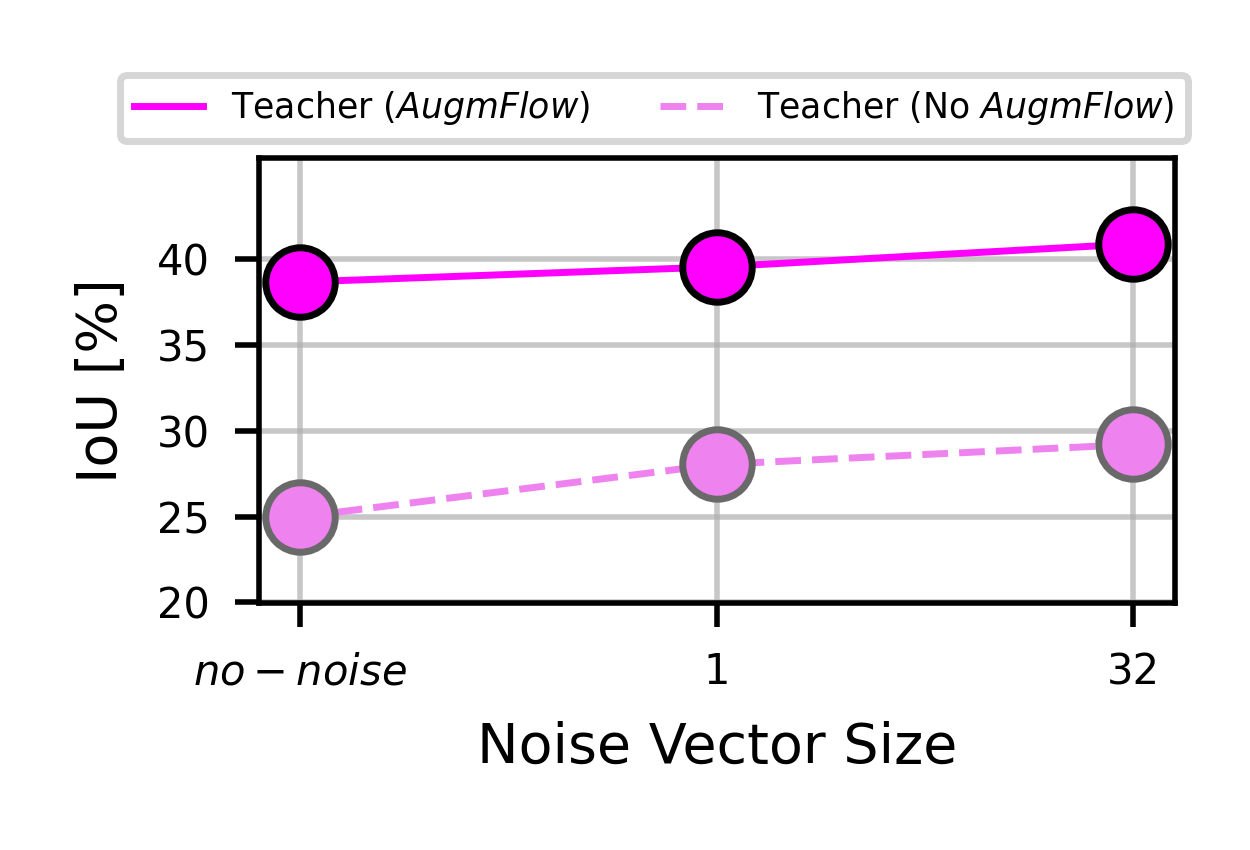}
      \caption{Analysis of the impact of noise vector size (no-noise, 1, 32) and flow augmentation $AugmFlow$ on optical-flow segmentation results by the \textit{Teacher} network on EndoVis2017VOS. Mean IoU [\%] is reported.} 
      \label{fig:cycle_noise}
\end{figure}
\begin{figure*}[!htpb]
    \centering
      \includegraphics[width=\textwidth]{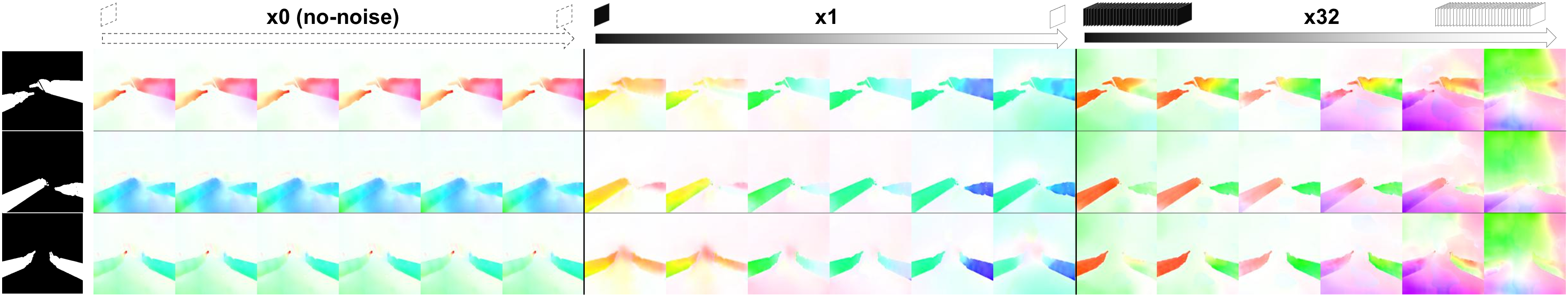}
        \caption{Qualitative results of the optical-flow generator ($G$), trained using different size of input noise vector among \{no-noise,1,32\}. First column: input \textit{shape-priors}; first block (x0), no noise concatenation; second block (x1), noise vector of size 1; third block (x32), noise vector of size 32. For each of the 3 blocks, from left to right, the noise vector was smoothly interpolated between all zeros to all ones (trivial for x0, having no concantenated noise).}.
      \label{fig:qual_gen_samples}
\end{figure*}

\subsection{Optical-Flow Augmentation and Noise Vector Size}
\label{sec:abl_flowaugm}
We first analyze optical-flow surgical tool segmentation by the \textit{Teacher} network. In particular, we evaluate the impact of the two proposed strategies to tackle the \textit{complexity-imbalance} between optical-flow and \textit{shape-priors} domain in the generative part of the \textit{Teacher} training, described in Section \ref{step1}: noise concatenation and optical-flow augmentation $AugmFlow$. We trained the \textit{Teacher} model using different sizes of the concatenated noise vector $n$, with and without the optical-flow augmentation $AugmFlow$.

Results shown in Figure \ref{fig:cycle_noise} highlight how optical-flow augmentation $AugmFlow$ plays a crucial role in counteracting \textit{complexity-imbalance}, allowing to reach quasi-optimal performance even without noise concatenation (continuous line, ``no-noise"). Noise concatenation also appears effective, with peak \textit{Teacher} performance reached with noise size 32 and $AugmFlow$. From qualitative results shown in Figure \ref{fig:qual_gen_samples}, it can be noted how noise concatenation allows to both generate more realistic and variable optical-flow images and disentangle tools configuration and optical-flow appearance. Note how, when changing \textit{shape-priors}, optical-flow image appearance changes when noise is not concatenated (x0, first block), but remains similar in case of noise concatenation (x1 and x32, second and third block, respectively).
It can also be observed how the most variable results are obtained with a noise vector size of 32 (third block, x32), with complexity increasing from leftmost column (noise vector of zeros, more frequently sampled during training) to rightmost column (noise vector of ones, rarely sampled during training), where tools are hardly recognizable.\\

\begin{figure}[t]
    \centering
      \includegraphics[width=3.in]{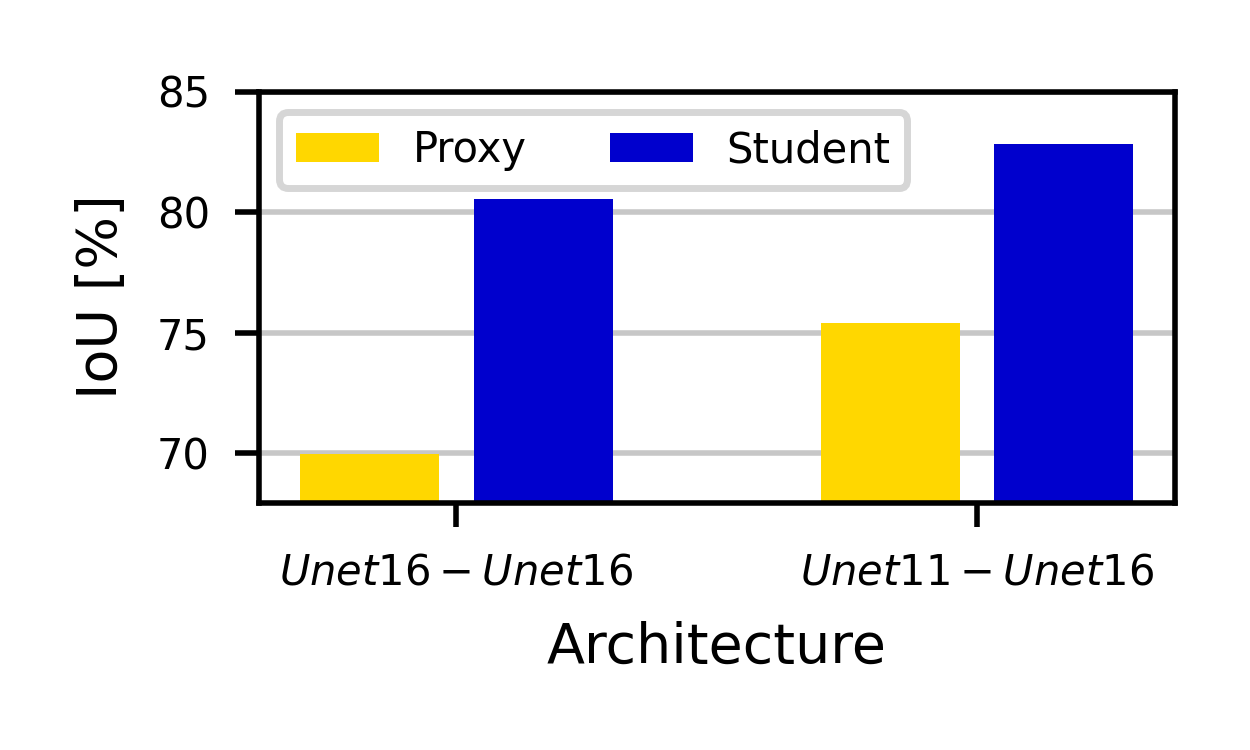}
      \caption{Analysis of the impact of \textit{Proxy} network's architecture on surgical tool segmentation results of \textit{Proxy} (yellow) and \textit{Student} (blue) networks, on EndoVis2017VOS. Mean IoU [\%] is reported.}       
      \label{fig:archit}
\end{figure}

\subsection{Proxy Network Architecture}
\label{sec:proxy_capacity}
In Section \ref{step2} we hypothesized the benefit of a limited \textit{Proxy} network capacity, in order to encourage the learning of the \textit{easiest} pattern shared between training samples, compatible with the pseudo-labels. We investigate this hypothesis by evaluating the performance of \textit{Proxy} and \textit{Student} networks, when using different \textit{Proxy} architectures (Unet11 and Unet16, defined in Section \ref{sec:impl_det}) on the EndoVis2017VOS dataset.

Results shown in Figure \ref{fig:archit} confirm that a shallower \textit{Proxy} network (Unet11) learns more effectively from the pseudo-labels than a deeper one (Unet16), quantified in an improvement of +5.44\% $\Delta$IoU. Additionally, this study provides the experimental proof that the \textit{Student}'s improvements with respect to the \textit{Proxy} are not due to their different architectures. Indeed, when using the same architecture for both of them (Figure \ref{fig:archit}, left) the \textit{Student} still outperforms the \textit{Proxy} by a large margin (+10.61\% $\Delta$IoU).

\begin{figure}[]
    \centering
      \includegraphics[width=3.in]{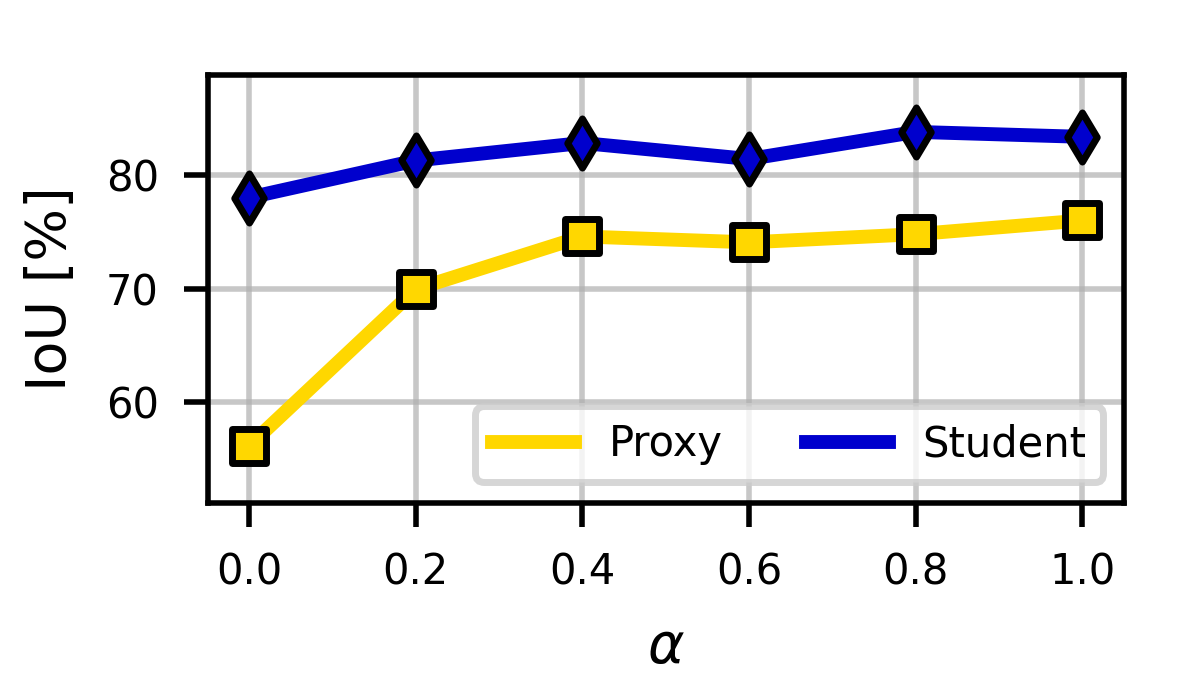}
      \caption{Analysis of the impact of loss function balancing coefficients $\alpha_{P}$ and $\alpha_{S}$ on  \textit{Proxy} (yellow) and \textit{Student} (blue) networks, on EndoVis2017VOS. We only consider the case $\alpha_{P}= \alpha_{S}=\alpha$; $\alpha$ equal 0 corresponds to cross-entropy loss only, $\alpha$ equal 1 corresponds to log IoU loss only. Mean IoU [\%] is reported.}      
      \label{fig:jaccard}
\end{figure}

\begin{figure*}[]
    \centering
      \includegraphics[width=\textwidth]{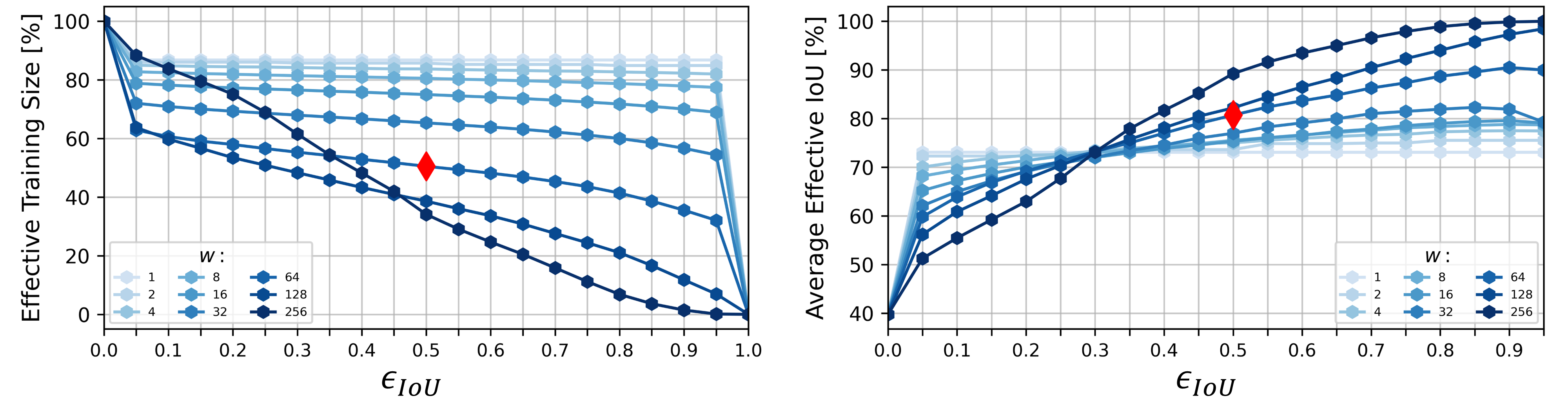}
      \caption{Impact of local IoU parameters ($\epsilon_{IoU}$ and window size $w$) on effective training size (left) and average effective IoU (right). x-axis can be interpreted as the level of agreement between \textit{Teacher} and \textit{Proxy} required in order to select a certain region (e.g. with $\epsilon_{IoU}$ equal to 0.8 a region is considered well-labelled only if the IoU between \textit{Proxy} and \textit{Teacher} predictions for that region is at least 80\%). Red markers correspond to $w=64$ and $\epsilon_{IoU}=0.5$, the values used in our main experiments.}
      \label{fig:eps_W}
\end{figure*}
\begin{figure}[]
    \centering
      \includegraphics[width=3.5in]{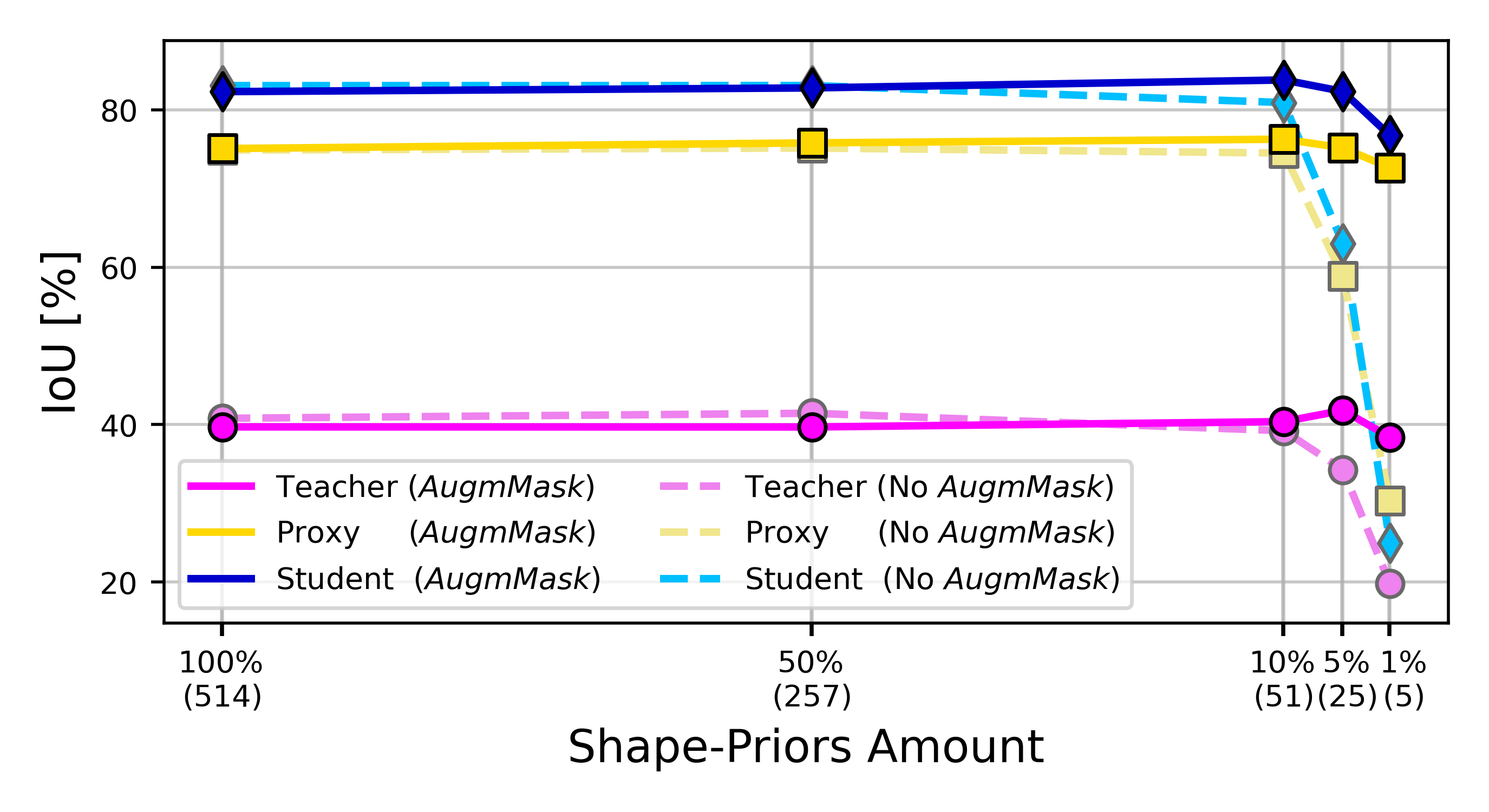}
      \caption{Analysis of the impact of decreasing \textit{shape-priors} quantity on individual frame and optical-flow segmentation, with and without $AugmMask$ augmentation, on EndoVis2017VOS. On the x-axis, the amount of RoboTool \textit{shape-priors} used for training is reported (absolute number and percentage with respect to the total number). Mean IoU [\%] for \textit{Student} (blue; dashed: trained without $AugmMask$), \textit{Proxy} (yellow; dashed: trained without $AugmMask$), \textit{Teacher} (purple; dashed: trained without $AugmMask$) is reported.}
      \label{fig:sp_size}
\end{figure}

\subsection{Loss Function Coefficients ($\alpha_{P}$, $\alpha_{S}$)}
\label{sec:jaccard_loss}
We investigate the impact of the balancing factors $\alpha_{P}$ and $\alpha_{S}$ between cross-entropy (CE) and log IoU losses in \textit{Proxy} and \textit{Student} networks training (Equations \ref{eq:loss_P}\&\ref{eq:loss_S}). In our experiments we consider the case $\alpha_{P}=\alpha_{S}=\alpha$, with $\alpha$
ranging from 0 (only CE loss) to 1 (only log IoU loss).

Results shown in Figure \ref{fig:jaccard} highlight the positive impact of log IoU loss, especially on the \textit{Proxy} network (+19.79\% $\Delta$IoU improvement between $\alpha$ = 1 and $\alpha$ = 0). This can be in part explained by the diminished-sensitivity of IoU based losses to class-imbalance. However, the greater improvement brought by the log IoU loss to the \textit{Proxy} network, directly trained on raw pseudo-labels, compared to the \textit{Student} network, may suggest that the log IoU loss is more robust to the noise of motion-derived pseudo-labels. Additional in-depth studies are required to investigate this hypothesis.

\subsection{Local IoU parameters' impact}
\label{sec:abl_epsw}
While state-of-the-art \textit{learning-from-noisy-labels} approaches usually require a \textit{Teacher} model trained on clean labels in order to identify well-labelled regions of noisy pseudo-labels, we perform this search in a fully-unsupervised way.
As detailed in Section \ref{step3}, \textit{probably} well-labelled regions are selected according to the \textit{agreement} between the pseudo-labels (\textit{Teacher} model's predictions from optical-flow segmentation $y_{t}^{T}$) and \textit{Proxy} model's predictions $y_{t}^{P}$. The agreement is measured by the \textit{local} IoU, parametrized by the window size $w$ ($w=h$ in our experiments), and binarized through the threshold parameter $\epsilon_{IoU}$, representing the minimum agreement required to consider a region well-labelled. The choice of these two parameters influences 1) the \textit{effective} number of pixels on which the \textit{Student} network is trained, 2) the average \textit{effective} IoU ($IoU_{eff}$) of the training labels, defined as the IoU between ground-truth masks GT and pseudo-labels $y_{t}^{T}$, computed only for the selected regions according to the binarized \textit{local} IoU ($\overline{IoU}_{(w,h)}^{loc}$) between $y_{t}^{T}$ and $y_{t}^{P}$:
\begin{equation}
    IoU_{eff} = \frac{|(GT\cap y_{t}^T) \cap \overline{IoU}_{(w,h)}^{loc}|}{|(GT\cup y_{t}^T) \cap \overline{IoU}_{(w,h)}^{loc}|}.
\end{equation}

We evaluate the influence of $w$ and $\epsilon_{IoU}$ on the \textit{effective} training size (expressed as total number of selected pixels over total number of pixels in the training dataset) and on the average $IoU_{eff}$ in the training dataset. For these experiments we considered trained \textit{Teacher} and \textit{Proxy} models on EndoVis2017VOS. We then varied $\epsilon_{IoU}$ and $w$ in a grid-like manner, with $\epsilon_{IoU}$ ranging from 0.0 to 1.0 with a step equal to 0.05, and $w$ in $\{1,2,4,8,16,32,64,128,256\}$. For each couple $(w,\epsilon_{IoU})$ we then evaluated \textit{effective} training size and average $IoU_{eff}$ on the EndoVis2017VOS training set, in order to provide an insight of the \textit{effective} training carried out.

Experimental results shown in Figure \ref{fig:eps_W} confirm that the agreement between pseudo-labels (optical-flow segmentation masks from the \textit{Teacher}) and \textit{Proxy} predictions is directly correlated to the quality of the pseudo-labels. Figure \ref{fig:eps_W}, right, shows the positive correlation between \textit{Proxy}-\textit{Teacher} agreement ($\epsilon_{IoU}$) and average \textit{effective} IoU, especially for large window sizes $w$ of the local IoU operation. As expected, the experiment also shows that requiring higher agreement reduces the amount of data effectively used for training, with a similar but inverse relationship. In light of this experiment, the values of window size $w$ and $\epsilon_{IoU}$ chosen for experimental validation, respectively 64 and 0.5, represent a good compromise, allowing to train the \textit{Student} network on 50.48\% of the total training data on EndoVis2017VOS, with an effective IoU of the pseudo-labels equal to 80.70\% (high-quality labels).\\

\subsection{Shape-Priors Quality \& Quantity}
\textit{Shape-priors} represent the only external information required by the proposed approach for training. In order to investigate their impact on the whole training process, we performed two sets of experiments. First, we evaluated the performance of our models (\textit{Teacher}, \textit{Proxy}, \textit{Student}) when trained using RoboTool and GrScreenTool \textit{shape-priors}, on EndoVis2017VOS, in order to evaluate the impact of different sources (i.e. \textit{recycled} annotations from a different dataset and automatically segmented tools from green-screen recordings); secondly, we trained our models using different percentages of the available RoboTool \textit{shape-priors}, from 100\% to 1\%, with and without on-the-fly augmentation $AugmMask$.\\
\begin{table}[t]
\centering
\begin{tabular}{|l|c|c|}
\hline
Shape-Priors & RoboTool & GrScreenTool \\
\hline
Teacher (ours)& 40.08 & 40.47\\
Proxy (ours)& 74.78 & 73.63 \\
Student (ours)& 83.77 & 82.63\\
\hline
\end{tabular}
\caption{Analysis of the impact of the \textit{shape-priors} dataset on frame segmentation. Comparison of the proposed method trained using RoboTool and GrScreenTool as \textit{shape-priors} on EndoVis2017VOS. Mean IoU [\%] is reported.}
\label{tab:abl_shapepriors}
\end{table} 
\begin{figure*}[!ht]
    \centering
      \includegraphics[width=\textwidth]{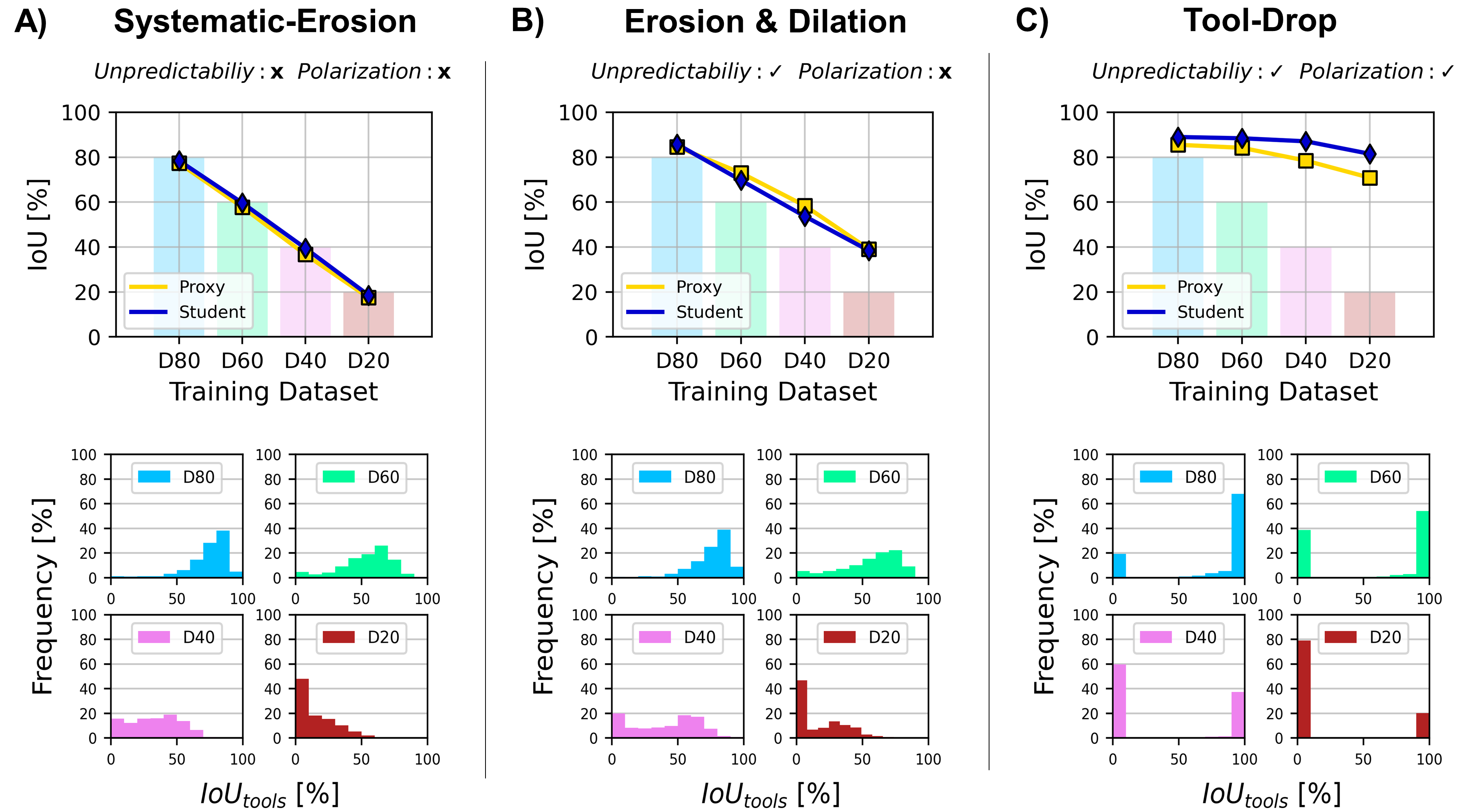}
      \caption{Analysis of the impact of \textit{unpredictability} and \textit{polarization} noise properties on the proposed method, on the artificially-corrupted EndoVis2017VOS datasets. Top: for each of the 3 noise sources (A, \textit{Systematic-Erosion}, predictable and not-polarized; B, random \textit{Erosion \& Dilation}, unpredictable and not-polarized; C \textit{Tool-Drop}, unpredictable and polarized) \textit{Proxy} (yellow) and \textit{Student} (blue) models were trained on the EndoVis2017VOS training dataset, having ground-truth labels corrupted with different levels of such noise. The colored bars are meant to improve readability, by visually showing the mean IoU between each training dataset labels and ground-truth clean labels ($\sim$80\% for D80, $\sim$60\% for D60, $\sim$40\% for D40, $\sim$20\% for D20);  Bottom: for each set of noisy labels, per-tool IoU histograms (IoU\textsubscript{tools}) computed as shown in Figure \ref{fig:polarization_exp}, are reported.}
      \label{fig:noise_exp}
\end{figure*} 

\begin{figure}[!ht]
    \centering
      \includegraphics[width=3.3in]{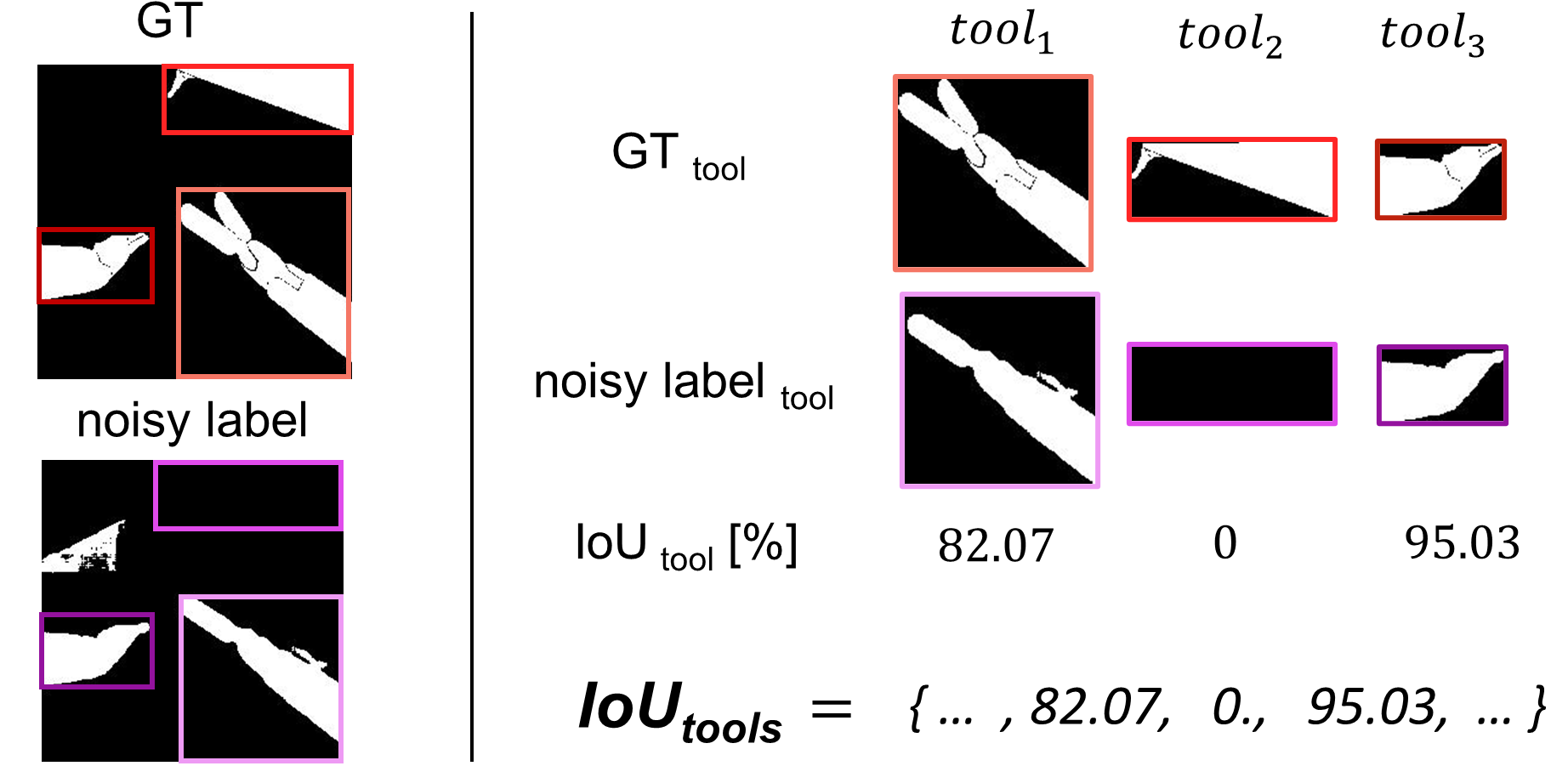}
      \caption{Computation of per-tool IoU between ground-truth masks and noisy labels. Left: example of ground-truth mask (GT) and noisy label. The smallest region containing each tool in the GT mask is extracted; the same exact region is extracted from the noisy label. Right: Intersection-over-Union (IoU\textsubscript{tool}) is computed between each region extracted from GT (GT\textsubscript{tool}) and noisy label (noisy label\textsubscript{tool}); the process is repeated for each tool in each frame of the dataset, and each IoU\textsubscript{tool} is stored in IoU\textsubscript{tools}. The distribution of per-tool IoU can then be visualized through histogram plots (Figures \ref{fig:noise_exp}\&\ref{fig:hit_teach}).}
      \label{fig:polarization_exp}
\end{figure}

\begin{figure}[!ht]
    \centering
      \includegraphics[width=2.8in]{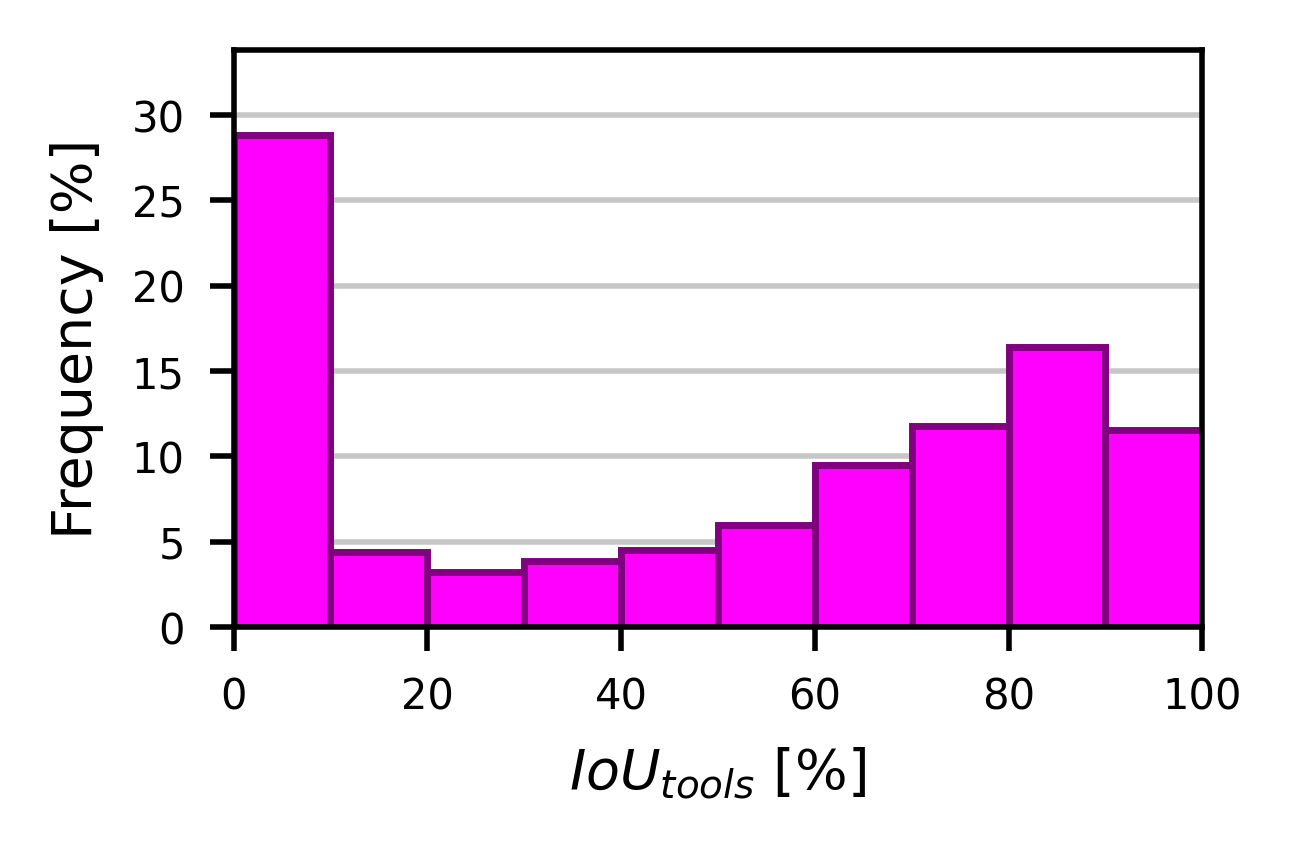}
      \caption{Per-tool IoU histogram (IoU\textsubscript{tools}), computed as shown in Figure \ref{fig:polarization_exp}, for pseudo-labels derived from motion segmentation by the \textit{Teacher} model on EndoVis2017VOS. Note how the distribution tends to be polarized on leftmost bin (completely mislabelled tools) and rightmost bins (\textit{almost}-perfectly segmented tools).}
      \label{fig:hit_teach}
\end{figure}

\indent Experimental results highlight how our FUN-SIS approach, although it requires \textit{shape-priors} as external source of information, has extremely loose requirements regarding their quality and quantity.  Experiments using GrScreenTool, reported in Table \ref{tab:abl_shapepriors}, provide comparable performance to the ones using RoboTool, despite the significantly different appearance of tools, as shown in Figure \ref{fig:shape_priors_ex}. In addition, experiments on \textit{shape-priors} quantity (Figure \ref{fig:sp_size}), show how the performance of \textit{Teacher}, \textit{Proxy} and \textit{Student} remains optimal even when using as few as 51 RoboTool \textit{shape-priors} masks (10\% of total) for training. If augmented on-the-fly using the $AugmMask$ protocol (random cropping and flipping), RoboTool \textit{shape-priors} can be further reduced to a total number of 5 instances (1\% of total), with limited performance drop (-5.57\% $\Delta$IoU compared to 100\% case).
\subsection{Noise properties (unpredictability \& polarization)}
\label{sec:abl_noise}

We investigate the impact of the \textit{unpredictability} and \textit{polarization} properties presented in Sections \ref{step2} and \ref{step3} on the proposed \textit{learning-from-noisy-labels} approach. To this aim, we carried out experiments with artificially controlled type and intensity of noise affecting the pseudo-labels, as described in Section \ref{sec:data_noisy}. We then substituted the pseudo-labels $y_{t}^{T}$, in our training pipeline,  with the corrupted EndoVis2017VOS labels and trained the \textit{Proxy} and the \textit{Teacher} networks according to the same modalities as the previous experiments. The three noise strategies presented in Section \ref{sec:data_noisy} were designed to highlight the effect of the \textit{unpredictability} and \textit{polarization} properties. In \textit{Systematic-Erosion} experiment, each mask was eroded, making the noise signal \textit{predictable} and \textit{not-polarized} (all tools are equally affected by the noise); in \textit{Erosion\&Dilation} experiment, each mask was either randomly eroded or dilated, making the noise signal \textit{unpredictable}, but still \textit{not-polarized} (each tool mask is affected by an error, either due to erosion or dilation); finally, in \textit{Tool-Drop} experiment, individual tools were either perfectly annotated or not annotated at all, making the noise signal both \textit{unpredictable} and \textit{polarized}.\\
\indent Results of the conducted experiments (Figure \ref{fig:noise_exp}) clearly highlight the impact of the two noise properties, as well as the ability of the proposed solution to leverage them.  When the noise is predictable (Figure \ref{fig:noise_exp}-A, top), the \textit{Proxy} network can perfectly learn to fit it, even when the corruption is minimal (D80). Contrarily, when noise cannot be inferred from single frames (Figure \ref{fig:noise_exp}-B\&C, top), the \textit{Proxy} network, unable to learn the noise pattern, will learn the easiest general pattern compatible with the labels,  resulting in significantly better predictions than the noisy labels used for its training (on average, +13.76\% $\Delta$IoU in \textit{Erosion\&Dilation}, +29.75\% $\Delta$IoU in \textit{Tool-Drop}). The effectiveness of the \textit{Student} network training is instead mainly influenced by the \textit{polarization} property. When the noise is not polarized (Figure \ref{fig:noise_exp}-A\&B, top), the \textit{Student} network does not benefit from region selection through \textit{local} IoU (+1.69\% and -1.87\% $\Delta$IoU, respectively, of \textit{Student} compared to \textit{Proxy} network). Instead, when the noise is polarized, well-labelled regions can be effectively identified using \textit{local} IoU, allowing for a consistent improvement of \textit{Student} predictions, compared to \textit{Proxy} ones (+6.73\% $\Delta$IoU on average, +8.60\% $\Delta$IoU in D40). The improvement is aligned with the one obtained in the experiments from Section \ref{sec:frame_eval} (+8.99\% $\Delta$IoU), where the pseudo-labels were produced via unsupervised surgical tool segmentation by the \textit{Teacher} network and had an IoU with the GT equal to 40.08\%. Overall, the proposed approach allows to maintain an IoU of at least 81.49\% (compared to the 88.99\% reached by fully-supervised training of the \textit{Student} model on clean labels, Table \ref{tab:all_end}), even when trained on extremely low-quality training labels (Figure \ref{fig:noise_exp}-C, top: \textit{Tool-Drop}, D20 i.e. $\sim$20\% IoU between training labels and GT). When trained on D80 and D60, the \textit{Student} network reaches optimal performance (88.98\% and 88.41\% IoU, respectively).\\
In order to provide a direct visualization of the \textit{polarization} property, we also report, for each set of noisy labels, including the motion-derived pseudo-labels by the \textit{Teacher} model, per-tool IoU histograms (IoU\textsubscript{tools}). Per-tool IoU can be computed, as shown in Figure \ref{fig:polarization_exp}, by extracting the smallest regions containing each tool from the GT labels, and computing the IoU between this region and the corresponding one from the corresponding pseudo-label. This process, while approximate (an extracted region from GT label may contain more than one tool), allows to produce a clear visualization of the \textit{polarization} property, by plotting the histogram of the obtained IoU\textsubscript{tools}. Histograms are shown in Figure \ref{fig:hit_teach}, for motion-derived pseudo-labels, and in Figure \ref{fig:noise_exp}, bottom, for artificially corrupted labels. From Figure \ref{fig:noise_exp}, bottom, it is possible to intuitively compare the case of not-polarized noise (A,B), where IoU\textsubscript{tools} values are mostly distributed around a single peak, to polarized noise (C), where the values appear concentrated on leftmost bin (full tool annotations missed) and rightmost bin (perfectly labelled tools). In the case of pseudo-labels derived from optical-flow segmentation (Figure \ref{fig:hit_teach}), the histogram, despite being smoothed by the sub-optimality of optical-flow estimator and segmenter described in Section \ref{step3}, still displays the \textit{polarization} property, allowing efficient \textit{Student} network training.

\subsection{Random Unlabelled Data}
In order to show the ease-of-use and robustness of the proposed FUN-SIS approach, we trained our models on the surgical robotic dataset RandSurg, described in Section \ref{sec:exp_modality} and tested on EndoVis2017VOS. The RandSurg dataset was created by collecting random public videos of surgical procedures, and performing minimal data curing. Training was carried out according to the same modalities as the other experiments, using RoboTool \textit{shape-priors} and varying amounts of the RandSurg data, ranging from very few (31 i.e. 1\% of total available) to all the available frames (3136).

\begin{figure}[tpb]
    \centering
      \includegraphics[width=3in]{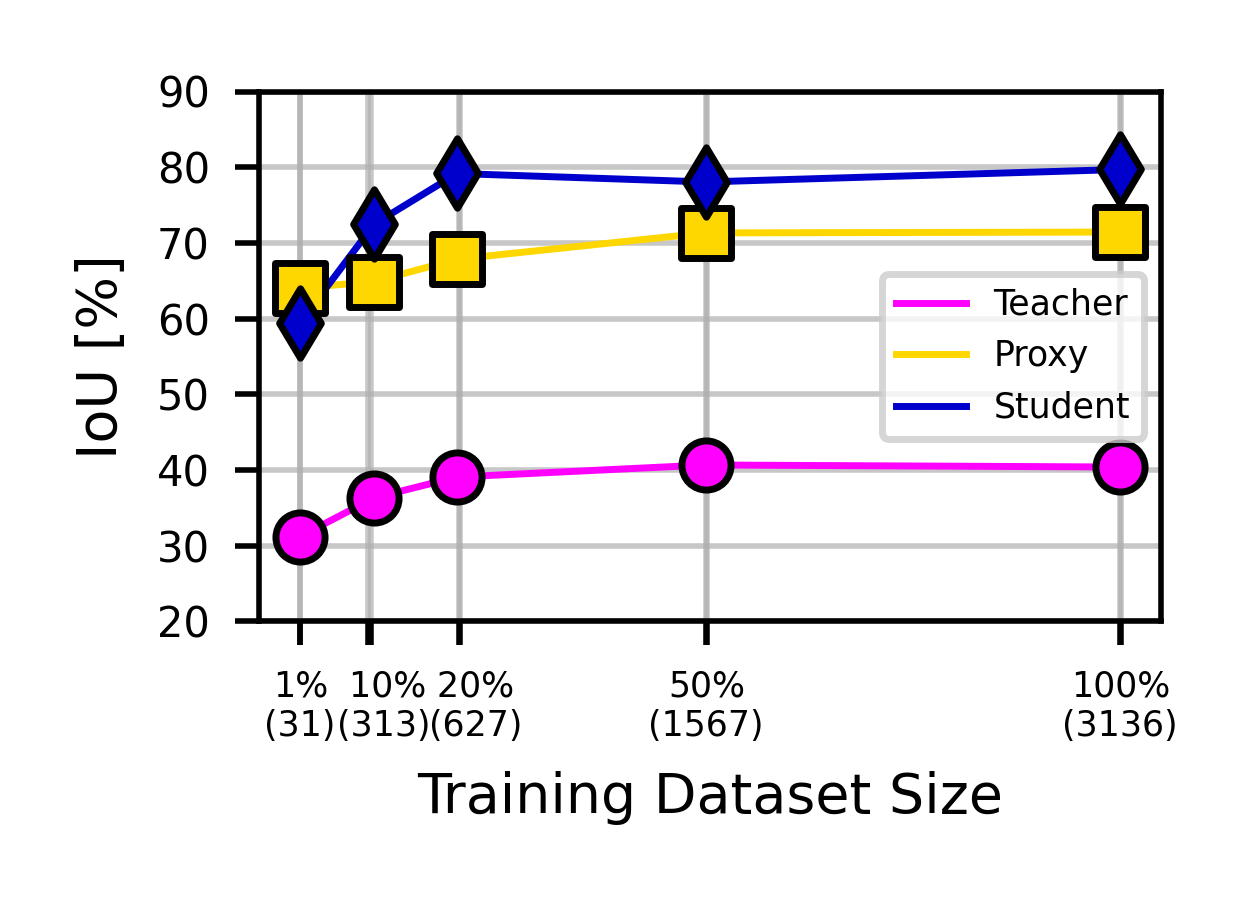}
      \caption{Analysis of proposed method performance when trained on increasing amounts of unlabelled RandSurg data, a dataset consisting of randomly selected surgical videos, downloaded from the public repository \cite{videos}, and tested on EndoVis2017VOS. On the x-axis, the amount of RandSurg frames used for training is reported (absolute number and percentage with respect to the total number). Mean IoU [\%] for \textit{Student} (blue), \textit{Proxy} (yellow), \textit{Teacher} (purple) is reported.}
      \label{fig:randsurg}
\end{figure}

Experimental results shown in Figure \ref{fig:randsurg} show that, despite the limited data curing and pre-processing of the input data, the method can easily leverage the increasing amount of available data to effectively train the models.  The \textit{Student} network reaches a peak IoU equal to 79.65\% on EndoVis2017VOS, comparable to the 83.77\% obtained when training on unlabelled data from the same dataset (Table \ref{tab:all_end}).

\subsection{FUN-SIS applicability on another domain: Cholec80}
We demonstrate the applicability of the proposed FUN-SIS approach on a different domain than the robotic one it was validated on. To this aim, we trained and qualitatively tested our \textit{Student} model on the unlabelled Cholec80 dataset, consisting of manual laparoscopic cholecystectomy procedures. Training was carried out using RoboTool \textit{shape-priors}, despite the different appearance of tools between robotic and manual laparoscopic videos.

\begin{figure*}[htpb]
    \centering
      \includegraphics[width=\textwidth]{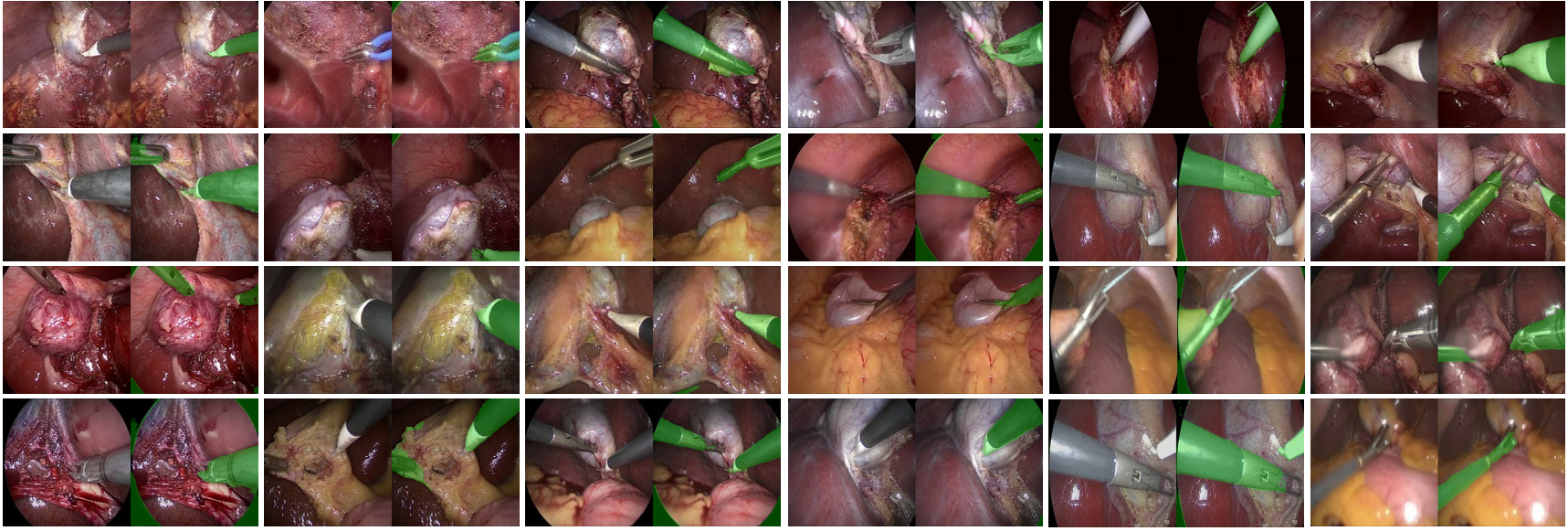}
        \caption{Qualitative results on the Cholec80 dataset. Original frame and overlapping between \textit{Student} network prediction and original frame are shown. Training was carried out using RoboTool \textit{shape-priors}.}           
        \label{fig:cholec80}
\end{figure*}

Results shown in Figure \ref{fig:cholec80} qualitatively confirm that the proposed method is applicable to a different surgical domain, even without domain-specific hyper-parameters tuning and with minimal pre-processing. Furthermore, they prove that despite the differences between \textit{shape-priors} and target tools, segmentation can still be effectively carried out. 

\section{Discussion and Future Work}
\label{sec:future}
In order to validate the proposed FUN-SIS approach, several experiments were performed and presented, including optical-flow segmentation (Section \ref{sec:exp_of}), per-frame segmentation (Section \ref{sec:frame_eval}, main experiment) and several ablation studies (Section \ref{sec:exp_abl}), dissecting the method and highlighting its key aspects.
The obtained results strongly support the soundness of FUN-SIS: binary surgical tool segmentation was effectively carried out in various datasets including EndoVis2017 (robotic surgery), STRAS (flexible endoscopic surgery), and Cholec80 (manual laparoscopic surgery).
When evaluated on EndoVis2017VOS, our \textit{Student} network reaches an IoU of 83.77\%, 12.30\% above the state-of-the-art unsupervised AGSD approach, and only 5.84\% below the state-of-the-art MF-TAPNet approach. 
Additionally, the proposed unsupervised approach for surgical tool segmentation of optical-flow images outperforms state-of-the-art approaches by a large margin on EndoVis2017VOS (+16.32\% $\Delta$IoU). Ablation studies proved that the method is extremely robust to the way \textit{shape-priors} are obtained, with no significant performance difference between using automatically segmented tools from green-screen recordings and \textit{recycled} manual annotations from other datasets. In addition, FUN-SIS showed great robustness to limited \textit{shape-priors} quantity, performing optimally on EndoVis2017VOS even using as few as 51 RoboTool \textit{shape-priors} masks for training. Ablation studies highlighted other interesting aspects, as the benefits of using a log Intersection-over-Union loss when training on noisy pseudo-labels, and the effectiveness of the proposed optical-flow augmentation strategy on video object segmentation. Finally, the extensive analysis on pseudo-label noise properties and their impact on neural-network training, as well as the proposed \textit{learning-from-noisy-labels} strategy to leverage them, may serve as base for future work on object segmentation using noisy labels, still largely unexplored.\\
Despite the satisfying results, the proposed work still presents potential room for improvement:
\begin{itemize}
    \item when selecting well-labelled regions through \textit{local} IoU, a great amount of the available data are currently discarded (49.52\% of total available pixels in EndoVis2017VOS experiment). These \textit{uncertainly}-labelled pixels could be exploited with semi-supervised-\textit{like} strategies, and contribute to the \textit{Student} network training;
    \item the window used to compute the \textit{local} IoU has fixed dimensions and is slid regularly on the masks with fixed width and stride; a more flexible approach, adapting to the varying tool size and location, may be beneficial;
    \item the \textit{Proxy} network is subjected to strong gradients while training directly on the noisy pseudo-labels, resulting in possible performance oscillations. This can potentially hinder the \textit{Student} network training, if the \textit{Proxy} network training is stopped in a poor weight parameters configuration. This problem could be mitigated by using approaches such as self-ensembling (\cite{nguyen2019self}), regularizing \textit{Proxy} network training;
    \item the FUN-SIS performance is overall influenced by the quality of the optical-flow images, which depends, in turn, on the endoscopic camera resolution and the optical-flow estimator. Current research on models for optical-flow computation specifically tailored for endoscopic images, as well as the increasing use of high-definition endoscopic cameras, could naturally contribute to improve the effectiveness of the proposed FUN-SIS method;
    \item the FUN-SIS approach, completely relying on instrument motion, is currently unable to perform semantic differentiation among the \textit{instrument} class. Strategies to extend FUN-SIS to multi-class segmentation could be explored, possibly involving motion patterns analysis and the use of limited external semantic supervision.
\\
    
\end{itemize}

\section{Conclusion}
In this paper we presented FUN-SIS, a novel Fully-UNsupervised approach for Surgical Instruments Segmentation. FUN-SIS effectively trains a per-frame surgical tool segmentation model on completely unlabelled endoscopic videos, solely relying on implicit motion information and instrument \textit{shape-priors}. In order to achieve this, we made several contributions, including a novel unsupervised optical-flow tool segmentation approach and a newly designed \textit{learning-from-noisy-labels} strategy. The proposed contributions were extensively validated on different surgical datasets (flexible endoscopic, robotic and laparoscopic procedures). On the popular MICCAI 2017 EndoVis Robotic Instrument Segmentation Challenge dataset, the proposed unsupervised approach performs almost on par with state-of-the-art fully-supervised models. \\
In conclusion, we hope that this work can contribute to the development of new segmentation methods requiring reduced supervision for training, fully exploiting the massive amounts of data which minimally invasive surgery can provide.

\section*{Acknowledgments}
This work was supported by the ATLAS project. The ATLAS project has received funding from the European Union’s Horizon 2020 research and innovation programme under the Marie Sklodowska-Curie grant agreement No. 813782. This work was also partially supported by French State Funds managed by the Agence Nationale de la Recherche (ANR) through the Investissements d’Avenir Program under Grant ANR-11-LABX-0004 (Labex CAMI) and Grant ANR-10-IAHU-02 (IHU-Strasbourg).

\bibliographystyle{model2-names.bst}\biboptions{authoryear}
\bibliography{references}

\clearpage

\appendix

\section{Implementation Details}
\label{sec:app_impl_det}
In our implementation, all the segmentation models have a U-Net-like architecture. The \textit{Teacher} network has a 5-convolutional-layers encoder (Figure \ref{fig_sup:Teacher}); the \textit{Proxy} network has a 11-convolutional-layers encoder (Figure \ref{fig_sup:Proxy}); the \textit{Student} network has the same architecture as TernausNet-16 (\cite{shvets2018automatic}), using a VGG-16 architecture as encoder, initialized from ImageNet pre-training. The generator network $G$ also has a U-Net-like architecture, but uses bilinear-upsampling instead of deconvolution in the expanding path (Figure \ref{fig_sup:gen}). The discriminator model is implemented using two separate neural networks, one producing a single score as output, another one producing a 16x16 local score-map, in charge of global and local appearance, respectively (Figure \ref{fig_sup:discr}).\\
Training parameters, determined from preliminary experiments on external data (\textit{phantom} dataset from \cite{sestini2021kinematic}), are reported in Table \ref{tab_sup:train_params}.

\setcounter{figure}{0}
\renewcommand{\thefigure}{A\arabic{figure}}
\renewcommand{\theHfigure}{A\arabic{figure}}
\setcounter{table}{0}
\renewcommand{\thetable}{A\arabic{table}}
\renewcommand{\theHtable}{A\arabic{table}}

\begin{figure}[h]
    \centering
      \includegraphics[width=3.5in]{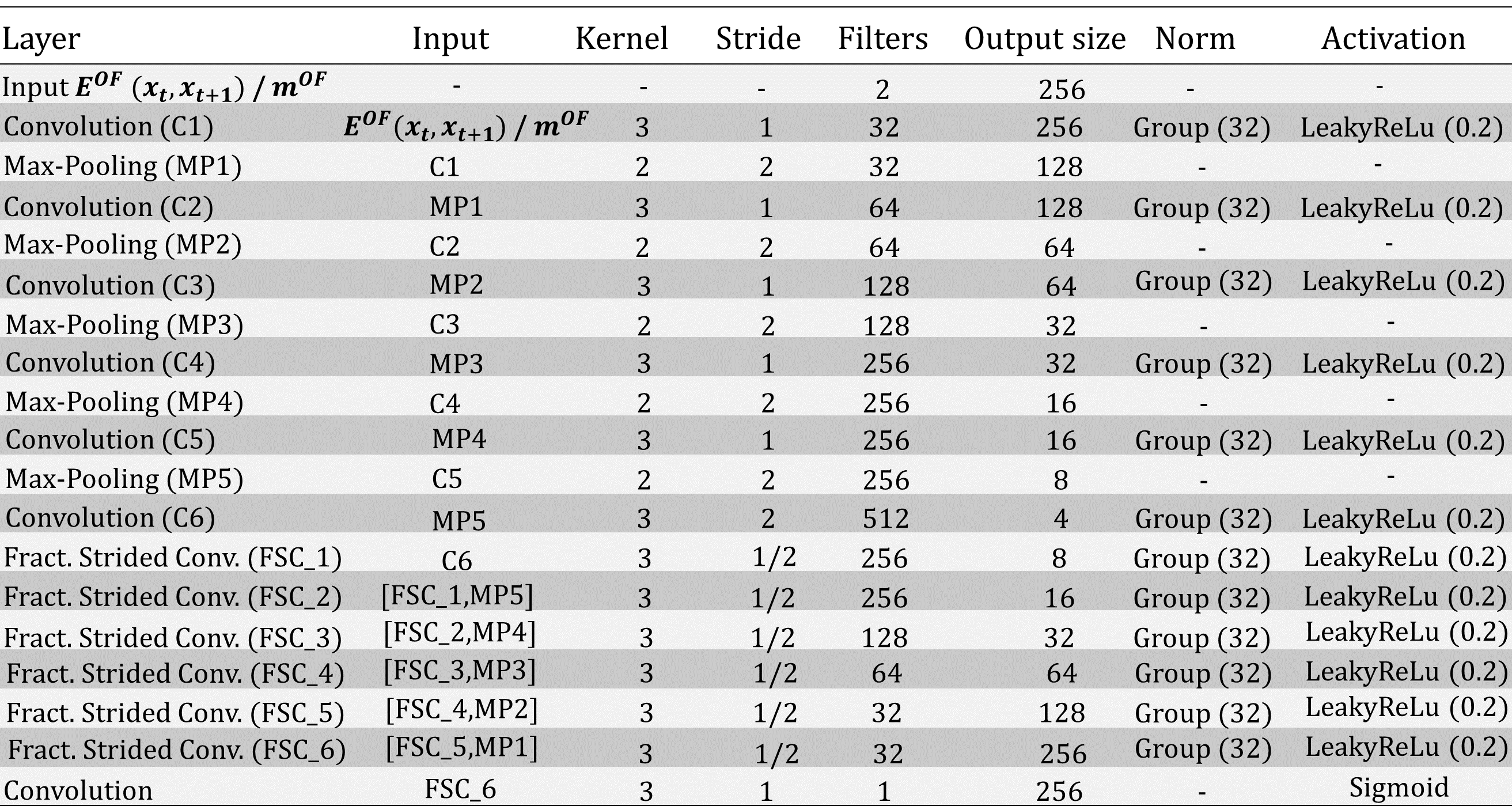}
      \caption{Network architecture of \textit{Teacher} optical-flow segmentation model.}      
      \label{fig_sup:Teacher}
\end{figure}

\begin{figure}[h]
    \centering
      \includegraphics[width=3.5in]{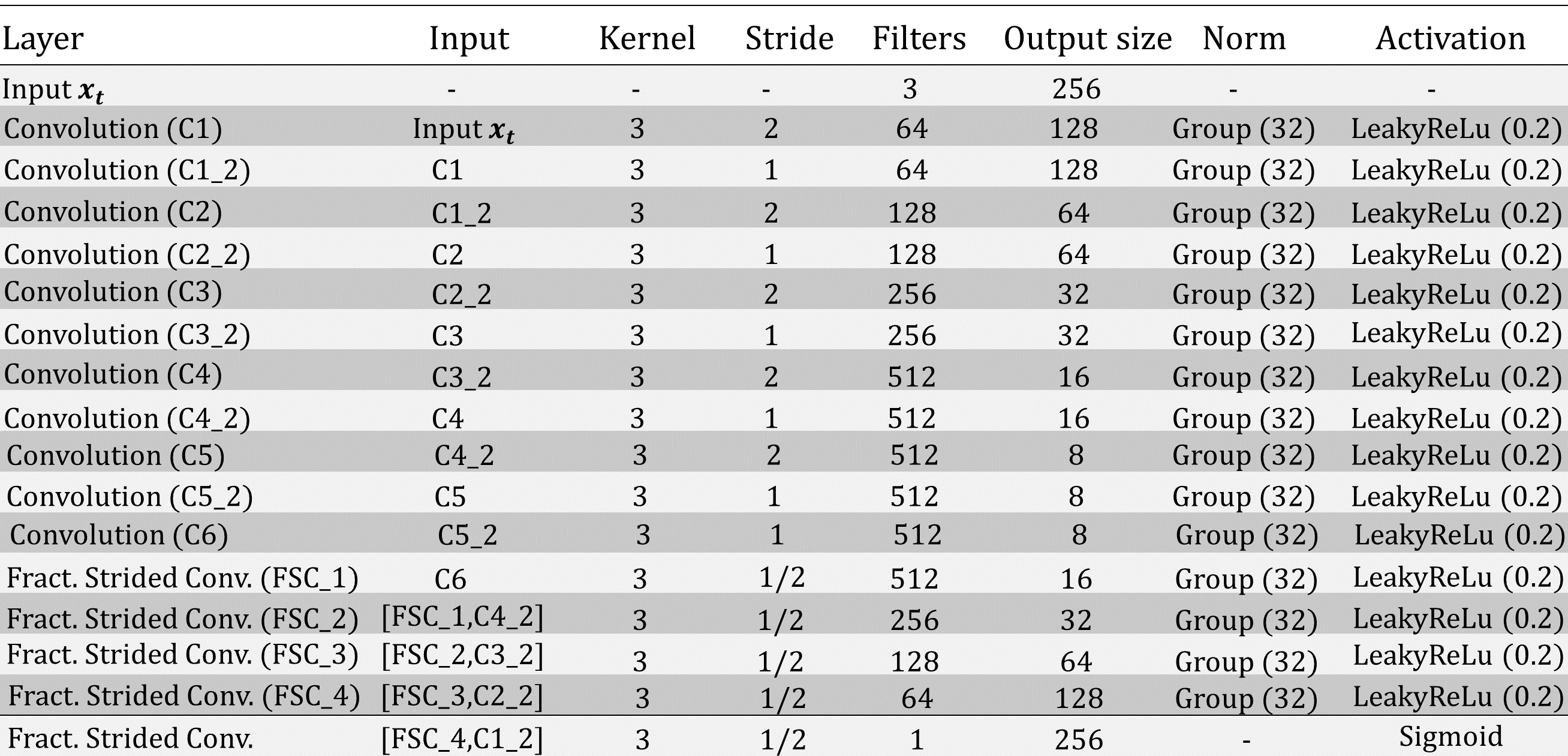}
      \caption{Network architecture of \textit{Proxy} segmentation model.}
      \label{fig_sup:Proxy}
\end{figure}

\begin{figure}[h]
    \centering
      \includegraphics[width=3.5in]{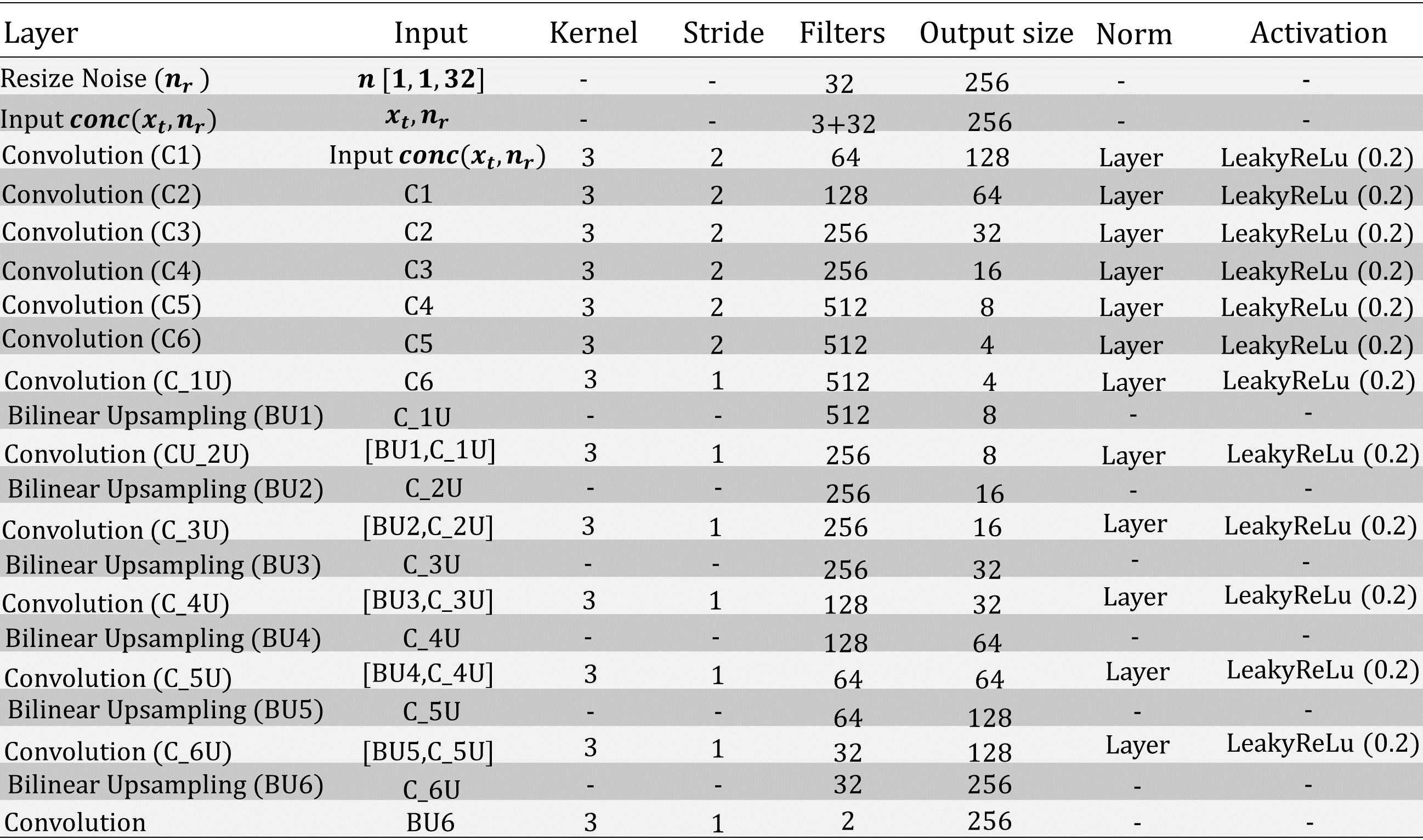}
      \caption{Network architecture of optical-flow generator model ($G$).}      
      \label{fig_sup:gen}
\end{figure}

\begin{figure}[h]
    \centering
      \includegraphics[width=3.5in]{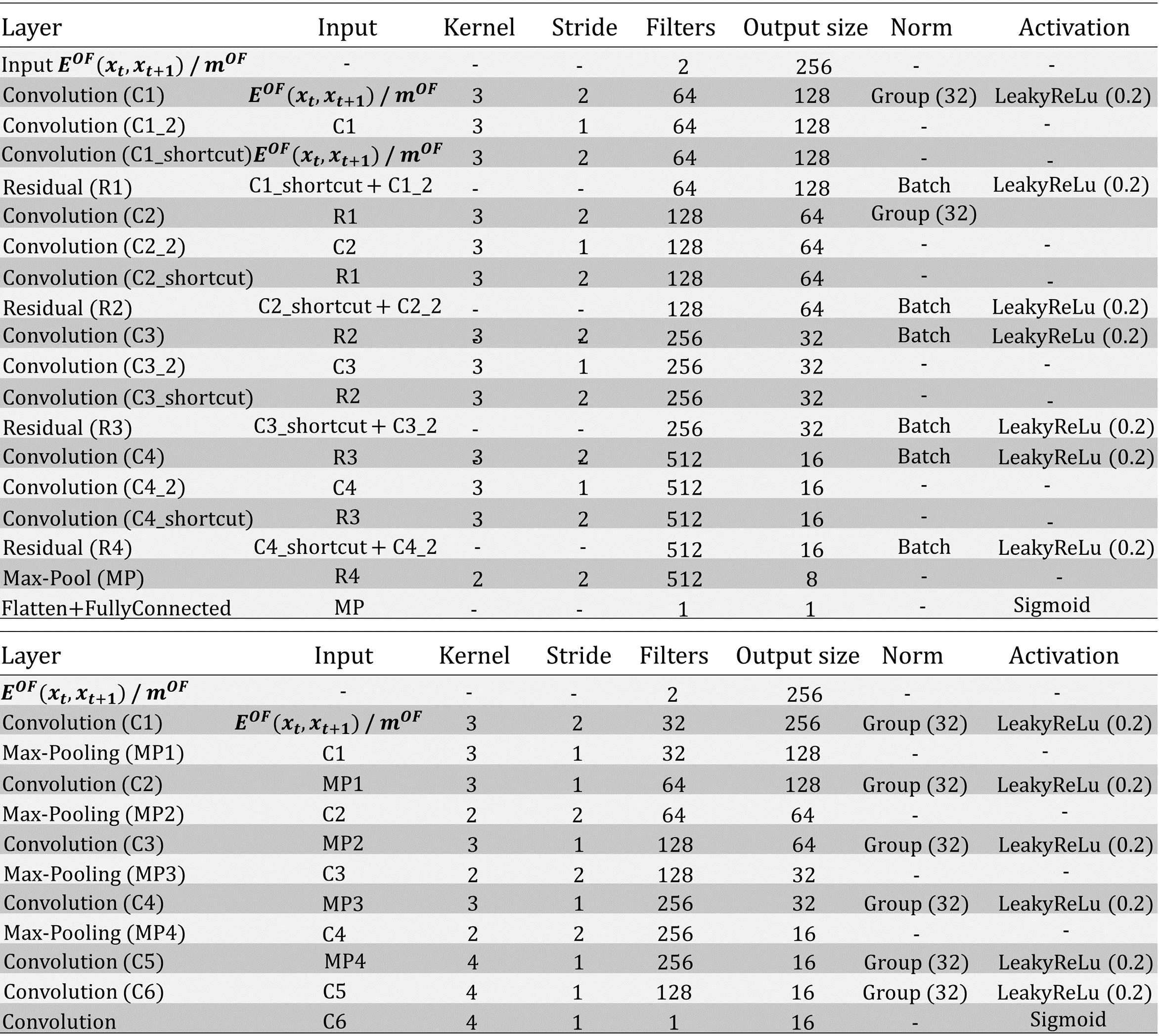}
      \caption{Network architecture of discriminator model $D$. Top: global discriminator, outputting a single global score; bottom: patch discriminator, outputting a 16x16 score-map.}      
      \label{fig_sup:discr}
\end{figure}

\begin{table}[!ht]
\centering
\begin{tabular}{|l|c|}
\hline
n\textsubscript{epochs} & 40\&40  \\
Batch size & 16  \\
LR\textsubscript{GAN} & \num{3e-3}\\
LR\textsubscript{Teacher} & \num{2e-3}\\
LR\textsubscript{Proxy} & \num{5e-4} ($\div2$ / 5 epochs, after epoch 20)\\
LR\textsubscript{Student} & \num{5e-5} ($\div2$ / 5 epochs, after epoch 20)\\
$\beta_1$\textsubscript{GAN} & 0.5\\
$\beta_2$\textsubscript{GAN} & 0.9\\
$\beta_1$\textsubscript{Teacher,Proxy,Student} & 0.9\\
$\beta_2$\textsubscript{Teacher,Proxy,Student} & 0.999\\
$\epsilon_T=\epsilon_P$ & 0.5\\
$\alpha_{P}=\alpha_{S}=\alpha$ &  0.8\\

\hline
\end{tabular}
\caption{Training hyper-parameters used in our experiments. Parameters reported: number of training epochs (n\textsubscript{epochs}) for step-1 (\textit{Teacher} and \textit{Proxy} training) \& step-2 (\textit{Student} training), batch size, learning-rates (LR), $\beta_1$ and $\beta_2$ for Adam optimizers, \textit{Teacher} and \textit{Proxy} binarization thresholds ($\epsilon_T$, $\epsilon_P$), loss balancing coefficients ($\alpha_P$, $\alpha_S$).}
\label{tab_sup:train_params}
\end{table}

\begin{figure*}[h]
    \centering
      \includegraphics[width=\textwidth]{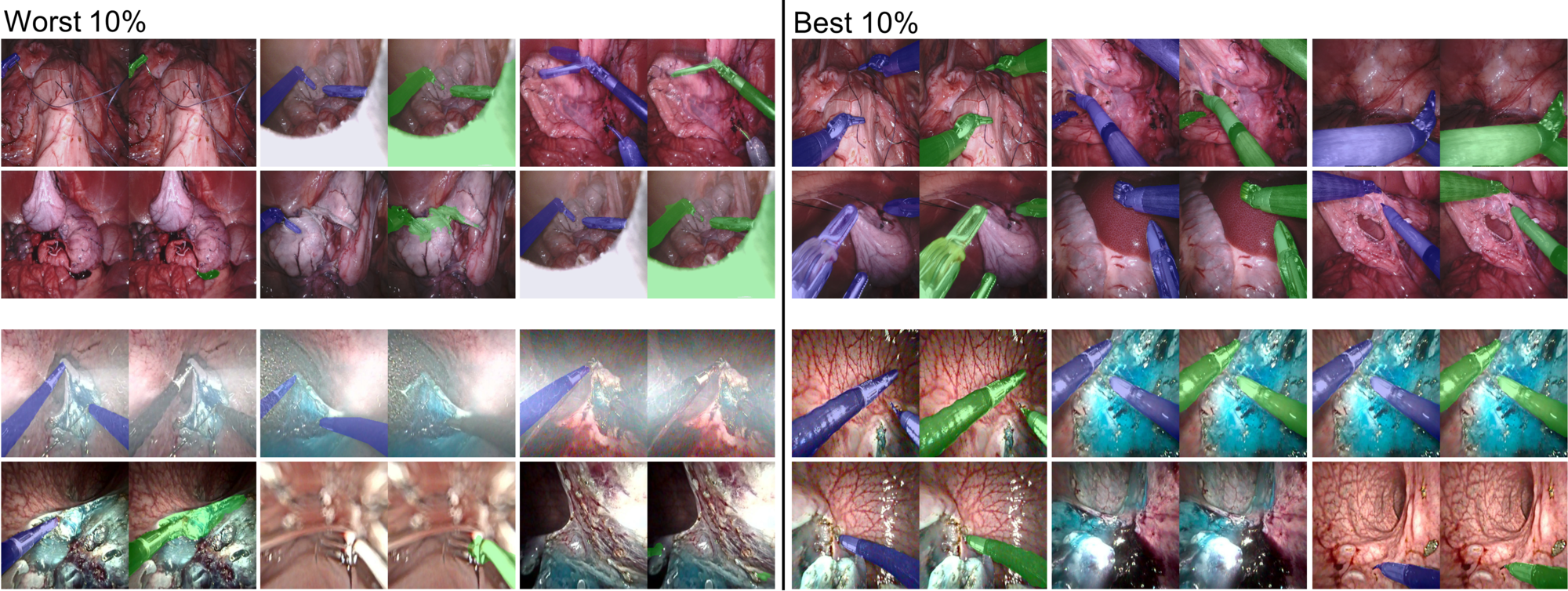}
      \caption{Qualitative results randomly drawn from worst (left) and best (right) 10\% predictions, according to IoU metric on EndoVis2017VOS (top) and STRAS (bottom) datasets. Original frame overlapped with ground-truth (blue) and \textit{Student} network's prediction (green).}      
      \label{fig_sup:bw}
\end{figure*}
\begin{figure*}[!h]
    \centering
      \includegraphics[width=\textwidth]{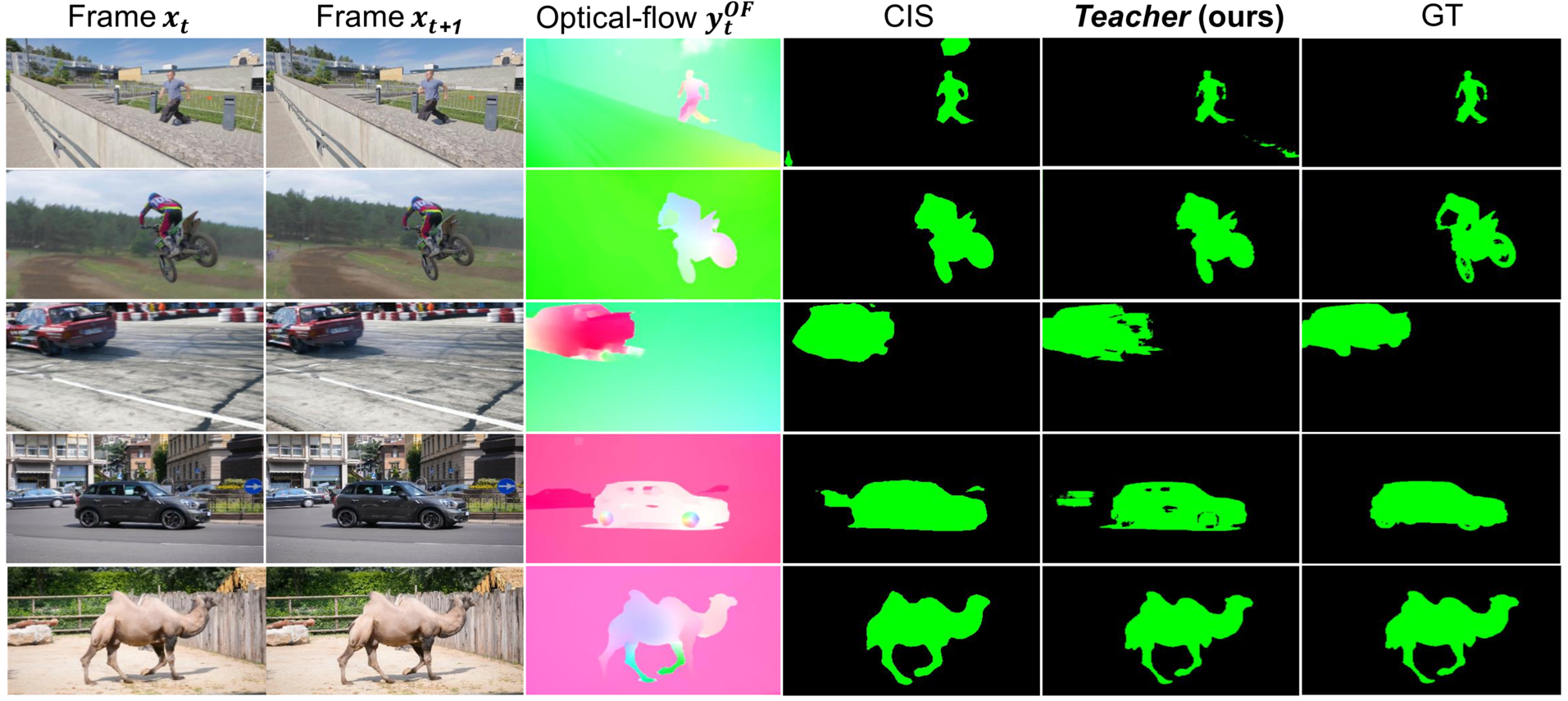}
      \caption{Optical-flow object segmentation on DAVIS2016 dataset. Qualitative results showing the two frames used for optical-flow computation, optical-flow image after HSV standard conversion, CIS (\cite{yang2019unsupervised}) and \textit{Teacher} (using SegTrackV2 \textit{shape-priors}) predictions, and ground-truth (GT).}      
      \label{fig_sup:qual_res_CIS}
\end{figure*}

\section{Additional Qualitative Results}
We report additional qualitative results for surgical tool segmentation experiment on EndoVis2017VOS and STRAS datasets (Tables \ref{tab:all_end}\&\ref{tab:all_stras} in the manuscript), randomly drawn from best and worst 10\% predictions of the experiments according to the IoU metric. Results are shown in Figure \ref{fig_sup:bw}.\\
We also report additional qualitative results for optical-flow object segmentation on DAVIS2016 dataset (Figure \ref{fig_sup:qual_res_CIS}) for the state-of-the-art CIS approach and our \textit{Teacher} model, trained using SegTrackV2 as \textit{shape-priors}.

\setcounter{figure}{0}
\renewcommand{\thefigure}{B\arabic{figure}}
\renewcommand{\theHfigure}{B\arabic{figure}}

\end{document}